\documentclass[sigconf]{acmart}
\usepackage{colortbl}
\usepackage[most]{tcolorbox}
\usepackage{booktabs}
\usepackage{color, colortbl}
\usepackage{multirow} 
\usepackage{array}    
\usepackage{amsmath}
\usepackage{diagbox}
\usepackage{array}
\usepackage{booktabs}
\usepackage{subcaption}
\usepackage{enumerate, enumitem}
\usepackage{makecell}
\usepackage[normalem]{ulem}
\usepackage{svg}
\usepackage{newfloat}
\usepackage{balance}
\usepackage{setspace}
\usepackage{amsmath,amsfonts,amsthm}
\usepackage{cleveref}
\usepackage{lipsum}
\usepackage{arydshln}
\usepackage{xcolor}
\usepackage{cleveref}
\crefname{section}{§}{§§}
\Crefname{section}{§}{§§}

\AtBeginDocument{%
  \providecommand\BibTeX{{%
    Bib\TeX}}}

\floatstyle{ruled}
\newfloat{listing}{tb}{lst}{}
\floatname{listing}{Listing}

\AtBeginDocument{%
  \providecommand\BibTeX{{%
    \normalfont B\kern-0.5em{\scshape i\kern-0.25em b}\kern-0.8em\TeX}}}

\crefname{section}{§}{§§}
\crefname{appendix}{Appendix}{Appendix}

\def\emb{\boldsymbol}

\definecolor{lightblue}{HTML}{D1E9F6}

\definecolor{high1}{RGB}{255, 170, 170}
\definecolor{high}{RGB}{197, 211, 232}
\definecolor{secondhigh}{RGB}{255, 227, 227}
\definecolor{gap}{RGB}{185, 229, 232}
\newcommand{\drp}{\cellcolor{gap}}

\newcommand{\fst}{\cellcolor{high1}}
\newcommand{\snd}{\cellcolor{secondhigh}}

\def\model{CSPO}
\newtheoremstyle{normalfont}
{5pt}   
{5pt}   
{\normalfont}  
{}      
{\bfseries} 
{.}      
{ }      
{}       
\theoremstyle{normalfont}

\newtheorem{definition}{Definition}

\newtheorem{problem}{Problem}

\DeclareMathOperator\soft{Softmax}

\copyrightyear{2025}
\acmYear{2025}
\setcopyright{acmlicensed}\acmConference[WWW Companion '25]{Companion Proceedings of the ACM Web Conference 2025}{April 28-May 2, 2025}{Sydney, NSW, Australia}
\acmBooktitle{Companion Proceedings of the ACM Web Conference 2025 (WWW Companion '25), April 28-May 2, 2025, Sydney, NSW, Australia}
\acmDOI{10.1145/3701716.3715216}
\acmISBN{979-8-4007-1331-6/25/04}
\begin{document}

\title{\model: Cross-Market Synergistic Stock Price Movement Forecasting with Pseudo-volatility Optimization}

\thanks{$^*$Equal Contribution. $^\dag$ Corresponding Authors.}

\author{Sida Lin$^*$$^\ddag$}
\affiliation{%
  \institution{The Chinese University of Hong Kong, Shenzhen}
  \thanks{$^\ddag$This work was conducted during his internship at IDEA Research.}
  \country{China}
}
\email{sidalin1@link.cuhk.edu.cn}

\author{Yankai Chen$^*$}
\affiliation{%
  \institution{Cornell University}
  \city{Ithaca}
  \country{United States}
}
\email{yankaichen@acm.org}

\author{Yiyan Qi}
\affiliation{%
  \institution{International Digital Economy Academy (IDEA)}
  \city{Shenzhen}
  \country{China}
}
\email{qiyiyan@idea.edu.cn}

\author{Chenhao Ma$^\dag$}
\affiliation{%
  \institution{The Chinese University of Hong Kong, Shenzhen}
  \country{China}
}
\email{machenhao@cuhk.edu.cn}

\author{Bokai Cao}
\affiliation{%
  \institution{The Hong Kong University of Science and Technology (Guangzhou)}
  \country{China}
}
\email{mabkcao@connect.hkust-gz.edu.cn}

\author{Yifei Zhang}
\affiliation{%
  \institution{Nanyang Technological University}
  \country{Singpore}
}
\email{yifei.zhang@ntu.edu.sg}

\author{Xue Liu}
\affiliation{%
  \institution{McGill University}
  \city{Montreal}
  \country{Canada}}
\email{xueliu@cs.mcgill.ca}

\author{Jian Guo$\dag$}
\affiliation{%
  \institution{International Digital Economy Academy (IDEA)}
  \city{Shenzhen}
  \country{China}
}
\email{guojian@idea.edu.cn}

%

\renewcommand{\shortauthors}{Sida Lin et al.}

\begin{abstract}

The stock market, as a cornerstone of the financial markets, places forecasting stock price movements at the forefront of challenges in quantitative finance.
Emerging learning-based approaches have made significant progress in capturing the intricate and ever-evolving data patterns of modern markets.
With the rapid expansion of the stock market, it presents two characteristics, i.e., \textit{stock exogeneity} and \textit{volatility heterogeneity}, that heighten the complexity of price forecasting.
Specifically, while stock exogeneity reflects the influence of external market factors on price movements, volatility heterogeneity showcases the varying difficulty in movement forecasting against price fluctuations.
In this work, we introduce the framework of \underline{C}ross-market \underline{S}ynergy with \underline{P}seudo-volatility \underline{O}ptimization (\model).
Specifically, \model~implements an effective deep neural architecture to leverage external futures knowledge. 
This enriches stock embeddings with cross-market insights and thus enhances the \model's predictive capability.
Furthermore, \model~incorporates \textit{pseudo-volatility} to model stock-specific forecasting confidence, enabling a dynamic adaptation of its optimization process to improve accuracy and robustness.
Our extensive experiments, encompassing industrial evaluation and public benchmarking, highlight \model's superior performance over existing methods and effectiveness of all proposed modules contained therein. 

\end{abstract}

\ccsdesc[500]{Computing methodologies~Neural networks}
\ccsdesc[500]{Mathematics of computing~Time series analysis}

\keywords{Stock price movement forecasting; Bayesian Neural Networks}

\maketitle
\section{Introduction}
Stock price movement forecasting, with the goal of predicting future upward/downward price trends, is a core task in quantitative investment with significant research attention~\cite{xu2021rest,zhao2023doubleadapt,lin2021learning}. 
Traditional approaches rely on manually constructed features from basic financial indicators, such as moving averages, price-to-earnings ratios, and trading volumes~\cite{box1970distribution,holt2004forecasting,kavitha2013stock}. 
While these features are with good interpretability, they may not capture the complex, dynamic, and nonlinear data patterns in markets. 

In recent years, machine learning and deep learning methods have revolutionized the area by enabling models to learn directly from raw financial data. 
Machine learning algorithms such as decision trees~\cite{quinlan1986induction,quinlan2014c4} and ensemble methods~\cite{zhou2012ensemble,chen2016xgboost,ke2017lightgbm,zhang2020doubleensemble} have been effectively applied to identify patterns with better robustness compared to traditional statistical methods. 
Deep learning models, particularly neural networks tailored for handling time-series financial data~\cite{du2021adarnn,wang2022adaptive,zhang2017stock}, have made great strides in capturing temporal dependencies. 
These methods are highly responsive to the dynamic nature of financial markets.
They have demonstrated strong performance in extracting localized price trend patterns~\cite{selvin2017stock,hoseinzade2019cnnpred}, thereby boosting the understanding of the market behaviors.

\begin{figure}[t]
    \centering
    \hspace{-0.5em}
    \includegraphics[width=0.48\textwidth]{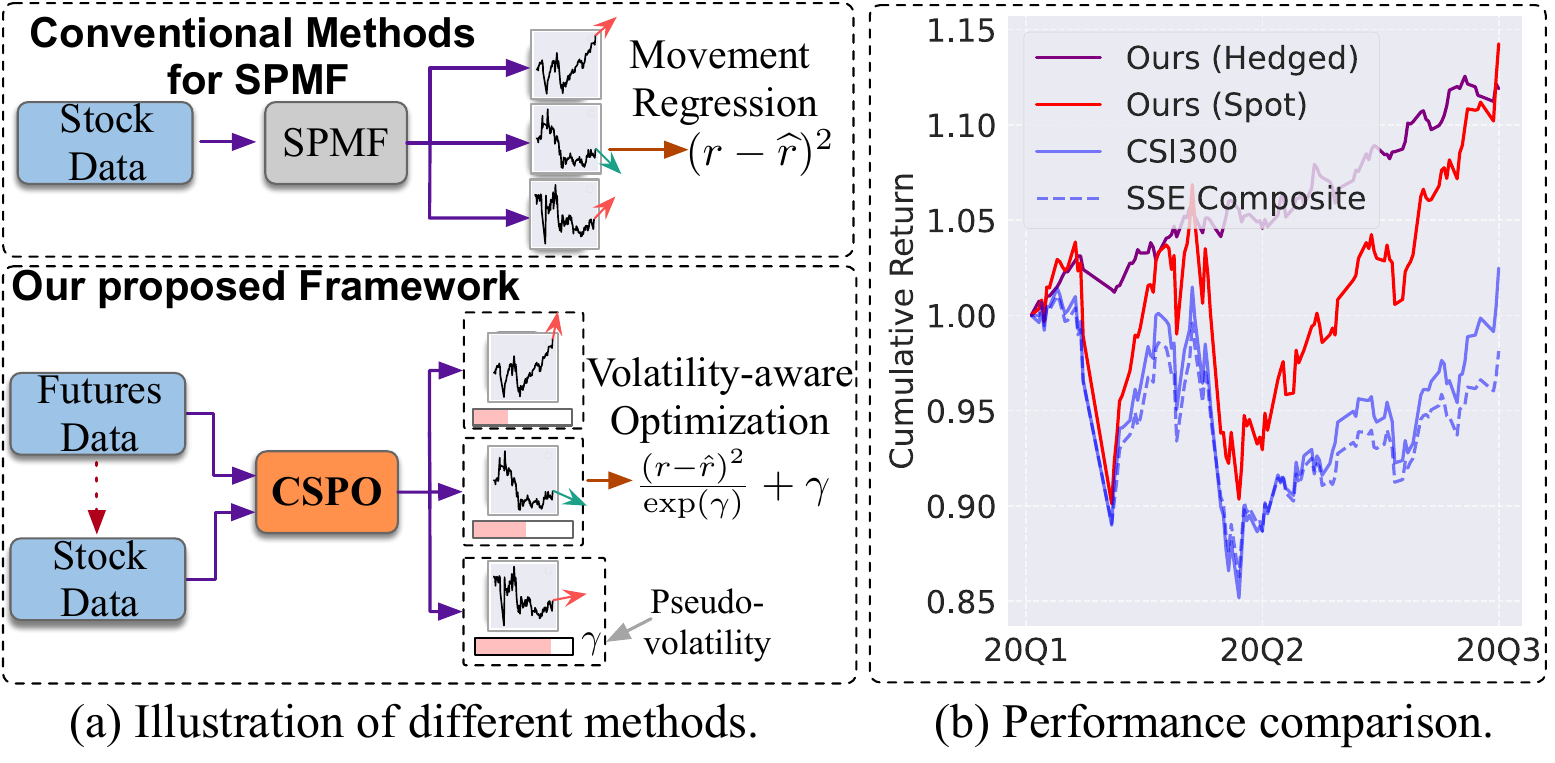}
    
    \caption{(a) Stock price movement forecasting (SPMF). (b) Portfolio yield comparison between baselines, i.e., CSI300 Index and SSE Composite, and our method to use \model~results as alpha-factors for hedged and spot return.}
    \label{fig:compare}
\end{figure}

As forecasting price movements appear to be increasingly difficult, the market has become more sophisticated and diverse, exhibiting two characteristics: \textit{stock exogeneity} and \textit{volatility heterogeneity}.
Specifically, (1) in the increasingly interconnected global economy, financial markets do not operate in isolation, where the stock market is affected by external events, e.g., policy changes, economic indicators, and geopolitical developments.
Beyond the conventional methods solely based on stock market data~\cite{zhang2017stock,lin2021learning}, recent deep forecasting models have started to integrate auxiliary information, such as exchanges~\cite{cao2012multifractal}, sales~\cite{zhang2022co}, and earnings calls~\cite{medya2022exploratory,yuan2023earnings,liu2024echo}, for performance improvement.  
This shows the promising potential of alleviating local limitations by incorporating spillover effects from other markets~\cite{lin2001spillover}.
(2) Different stocks with varying company-specific fundamentals respond differently to market stimuli, exhibiting varied volatility patterns.
Since volatility indicates the stability of stock price fluctuations, higher volatility thus increases the difficulty of price movement prediction.
Although only a few methods~\cite{zhu2022wise,li2024finreport} avoid assuming stock homogeneity, they primarily consider volatility to adjust the predicted price.
However, we argue that such volatility essentially presents the prediction confidence, not just adjustments to price values.
This approach may still overlook potential prediction deviations in model optimization, underscoring the need for new non-uniform learning strategies tailored to volatility heterogeneity.

Aligning with the aforementioned characteristics, we push forward the study of price movement forecasting by proposing \underline{C}ross-market \underline{S}ynergy with \underline{P}seudo-volatility \underline{O}ptimization framework (\model).
We provide a high-level illustration of \model~in Figure~\ref{fig:compare}(a).
(1) \uline{Firstly, to explore stock exogeneity, we propose incorporating \textit{futures market} information for forecasting guidance.}
Unlike other additional information sources, futures markets are inherently forward-looking, as they represent contracts to buy or sell assets on future dates. 
This provides practical insights into market expectations about future movements, which can be directly relevant for forecasting stock prices.
To achieve this, we design a transformer-based deep neural architecture, namely \textit{Bi-level Dense Pricing Transformer} (\texttt{BDP-Former}).
As the name suggests, \texttt{BDP-Former} progressively conducts cross-market knowledge synergy followed by price movement prediction.
This emulates real-world trading practices, where traders consider futures-to-stock and stock-to-stock correlations in their historical data and future trends~\cite{chui2012extreme}.
Therefore, it leverages a broader spectrum of market data to capture the complex interdependencies and enhances the model's predictive capability accordingly.
(2) \uline{Secondly, for volatility heterogeneity, we propose to study it within the model optimization process.}
In practice, traders consider not only the macro market factors but also the stability of individual stock prices.
Intuitively, volatility reflects price stability and thus indicates the confidence level in price forecasting. This motivates us to differentiate the loss contributions during optimization to minimize errors associated with low confidence predictions. 
To adapt to the time-varying and stock-specific volatility, we introduce the concept of \textit{pseudo-volatility} that can be estimated alongside our \model~framework.
Then the estimated pseudo-volatility is incorporated into our final objective function.
Compared to the conventional loss design, e.g., mean squared error, where they assume the equal loss contribution, 
our objective explicitly accounts for the loss variance inherent in different stock price predictions. 
This approach provides a more fine-grained learning optimization process, ultimately leading to more accurate price movement forecasting.

To comprehensively evaluate the performance of \model, we conduct extensive experiments on several real-world stock market datasets.
We first evaluate it in the industrial setting, where we leverage detailed proprietary backtesting with different evaluation strategies and metrics.
As shown in Figure~\ref{fig:compare}(b), our methods achieve more satisfactory yield curves on the CSI300 Index, i.e., a widely evaluated stock market dataset. 
We also provide a detailed public benchmarking with several existing models and empirical analyses of \model. 
The results further demonstrate not only the superiority of our framework over baselines but also the effectiveness of all proposed module designs contained within. 
To summarize, we have made the following threefold contributions:
\begin{itemize}[leftmargin=*] 
    \item We incorporate external futures information, and propose a deep neural architecture \texttt{BDP-Former} for effective cross-market knowledge synergy and enhanced stock price movement forecasting. 

    \item We introduce the estimation of \textit{pseudo-volatility} to capture price movement stability and forecasting confidence, which is further leveraged to differentiate their diversities in model optimization. 

    \item Extensive experiments on both industrial evaluation and public benchmarking demonstrate the effectiveness of our proposed framework \model.  
\end{itemize}



\begin{figure*}[t]
\begin{minipage}{\textwidth}
\centering
\includegraphics[width=0.9\linewidth, scale=0.8]{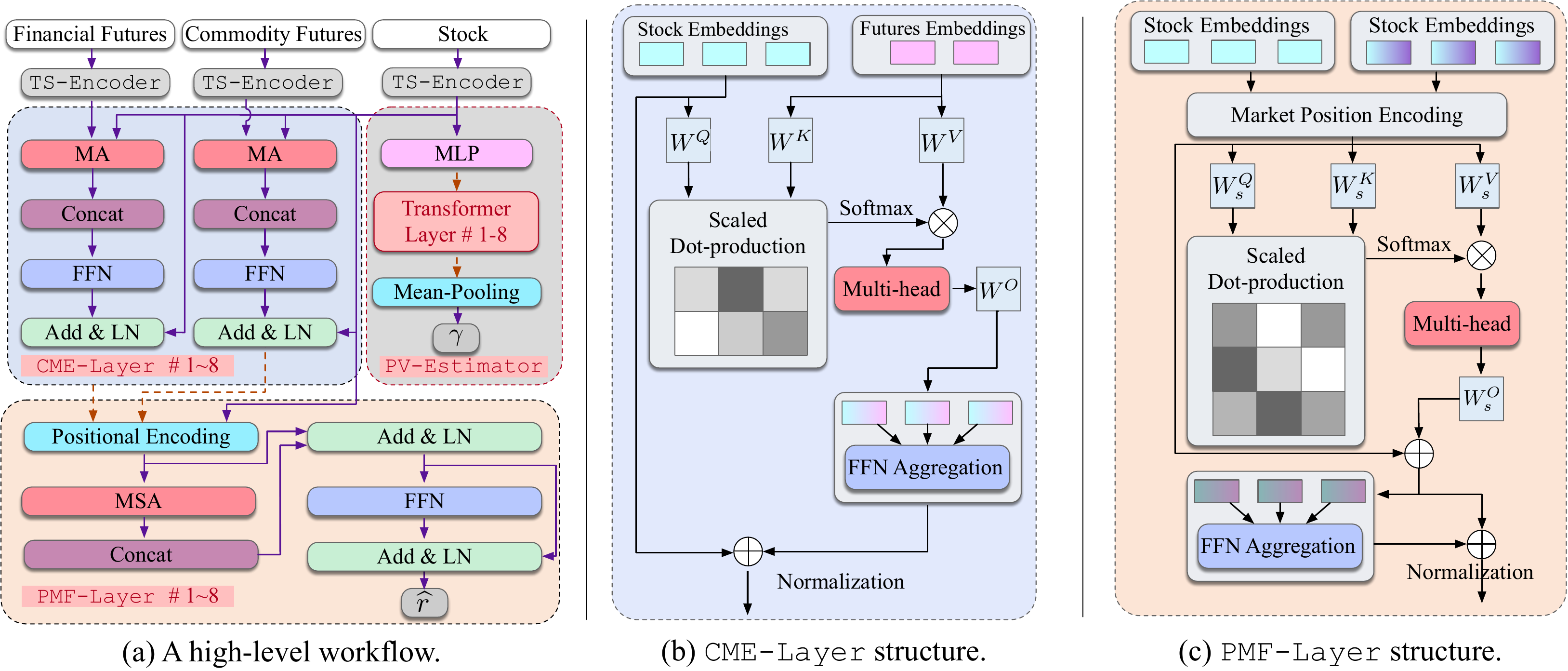}
\end{minipage} 
\caption{An illustration of our framework overview (best view in color). In \texttt{CME-Layer}, the notations, e.g., $\emb{W}^Q$, generalize to both cases of commodity futures and financial futures, e.g., $\emb{W}_c^Q$ and $\emb{W}_e^Q$.}
\label{fig:framework}
\end{figure*}

\section{Problem Formulation}
\label{sec:problem}
Assume that a stock market consists of $k_s$ stock assets. Let $t$ be the look-back window of the historical data.
At different time steps, each stock asset exhibits unique states, such as prices, market shares, etc. 
Thus we use $d^{'}$ features to characterize each asset.
We first introduce the data format for the stock market time series:
\begin{definition}[\textbf{Stock Market Time Series}]
Let $\emb{S} \in \mathbb{R}^{t \times k_s \times  d^{'}}$ denote 
the 3-dimensional tensor representing the stock market time series data.
$\emb{T}_i^{s}$ denotes the market snapshot with all stocks at the time step $i$.
$\emb{S}$ is formally defined as:
\begin{equation}
    \emb{S} = \big[\emb{T}_1^{s}, \emb{T}_2^{s}, \dots, \emb{T}_t^{s}\big], \text{ where } \emb{T}_i^{s} \in \mathbb{R}^{k_s \times  d^{'}}.
\end{equation}  
\end{definition}

For the futures markets, we consider both commodity and financial futures data:
\begin{definition}[\textbf{Futures Market Time Series}]
Let $\emb{C} \in \mathbb{R}^{t \times k_c \times  d^{'}}$ and $\emb{E} \in \mathbb{R}^{t \times k_e \times  d^{'}}$ respectively represent the \textit{commodity futures} and \textit{financial futures} time series.
$\emb{T}_i^{c}$ and $\emb{T}_i^{e}$ represent the snapshot of $k_c$ commodity futures and $k_e$ financial futures at time step $i$.
These futures market time series are formulated as:
\begin{equation}
\begin{aligned}
\emb{C} & = \big[{\emb{T}_1^{c}, \emb{T}_2^{c}, \dots, \emb{T}_t^{c}}\big], \text{ where } \emb{T}_i^{c} \in \mathbb{R}^{k_c \times d^{'}}, \\
\emb{E} &= \big[{\emb{T}_1^{e}, \emb{T}_2^{e}, \dots, \emb{T}_t^{e}}\big], \text{ where }  \emb{T}_i^{e} \in \mathbb{R}^{k_e \times d^{'}}.
\end{aligned}
\end{equation}
\end{definition}

The problem addressed in this paper is defined as follows:
\begin{problem}[\textbf{Stock Price Movement Forecasting}]
We aim to construct a deep neural predictive model, $f$, inputting $t$-size look-back window of data for stock market $\emb{S}$, commodity futures market $\emb{C}$, and financial futures market $\emb{E}$, and then predict the stock price movements, i.e., $\widehat{\emb{r}}_{t+1} \in \mathbb{R}^{k_s}$, on the market at the next time step, i.e., $\widehat{\emb{r}}_{t+1} = f_{1:t}(\emb{S, C, E})$.
Please notice that, $\widehat{\emb{r}}_{t+1}$ could be positive or negative; for stock prices at the $t$-th time step $\emb{P}_{t} \in \mathbb{R}^{k_s}$, $\widehat{\emb{r}}_{t+1} = \frac{\emb{P}_{t+1}}{\emb{P}_{t}} - \mathbf{1}$, where $\mathbf{1} \in \mathbb{R}^{k_s}$. 
We use $\widehat{\emb{r}}$ to refer to $\widehat{\emb{r}}_{t+1}$ for brevity whenever unambiguous.

\end{problem}

\section{\model~ Framework}
\label{sec:method}
\subsection{Overview}
In complex financial markets, futures and stocks are often interrelated, with mutual influences on pricing; our approach aims to capture these relationships for enhanced stock price movement forecasting.
Firstly, futures and stock time series data are encoded, as described in~\cref{sec:ts_encoding}. 
Then in \cref{sec:bdp}, we propose the \textit{Bi-level Dense Pricing Transformer} architecture to capture both futures-stock and stock-stock correlations for stock price movement forecasting. 
In \cref{sec:volatility}, we propose the \textit{pseudo-volatility} to associate with stock price prediction confidence, which improves model optimization through our volatility-aware objective function. 
The overall framework is illustrated in Figure~\ref{fig:framework}.

\subsection{\textbf{Market Time Series Encoding}}
\label{sec:ts_encoding}
For the input market time series data, e.g., stocks $\emb{S}$, commodity futures $\emb{C}$, and financial futures $\emb{E}$, 
a standard procedure is to encode them with $d$-dimensional representations for further model learning.
Therefore, we follow recent stock price prediction models~\cite{wang2022adaptive,xia_ci-sthpan_2024,gao_stockformer_2023} to adopt the encoder, denoted by \texttt{TS-Encoder}, for market information encoding:
\begin{equation}
\begin{aligned}
    \emb{S}^* & = \texttt{TS-Encoder}(\emb{S}), \text{ where } \emb{S}^* \in \mathbb{R}^{k_s \times d}, \\
    \emb{C}^* & = \texttt{TS-Encoder}(\emb{C}), \text{ where } \emb{C}^* \in \mathbb{R}^{k_c \times d}, \\
    \emb{E}^* & = \texttt{TS-Encoder}(\emb{E}), \text{ where } \emb{E}^* \in \mathbb{R}^{k_e \times d}.
\end{aligned}
\end{equation}

\subsection{Bi-level Dense Pricing Transformer}
\label{sec:bdp}
In this work, we draw inspiration from real-world practices where traders usually aggregate multiple sources of market information for stock future pricing.
We thus introduce our \textit{Bi-level Dense Pricing Transformer} structure, i.e., \texttt{BDP-Former}, for effective market information fusion and stock price forecasting:
\begin{equation}
\widehat{\emb{r}} = \texttt{BDP-Former}(\emb{S}^*, \emb{C}^*, \emb{E}^*).
\end{equation}
\texttt{BDP-Former} mainly consists of two levels of stacked layers, i.e., \textit{Cross-market Embedding Layer} and \textit{Price Movements Forecasting Layer}.
Generally, the first level of the learning layer captures latent relationships between futures and stocks, propagating futures-stock information to enrich stock embeddings with cross-market positional knowledge.
The second level then takes these aggregated stock representations as input to model stock-stock correlations, which ultimately enables stock price movement forecasting.

\subsubsection{\textbf{Cross-market Embedding Layer}}
In practice, futures and stock markets possess interconnections that share correlations in historical information and may impact future price movements.
To leverage these futures-stock connections, we introduce a dense connectivity module between futures and stocks, designed to aggregate diverse information from futures markets into stock embeddings.
In our work, it is stacked by 8 layers of \textit{Cross-market Embedding Layer} (\texttt{CME-Layer}) as follows:
\begin{equation}
\label{eq:cme_c}
    \overline{\emb{S}}_c = \texttt{CME-Layer}^{[8]}(\emb{S}^*, \emb{C}^*), \text{ where } \overline{\emb{S}}_c \in \mathbb{R}^{ k_s \times d}.
\end{equation}
Here stock embeddings are derived from the commodity futures.
We use commodity futures as the example for following explanations.

To implement \texttt{CME-Layer}, we leverage the Self-attention (SA) mechanism~\cite{vaswani2017attention} followed by the simplified forward networks.
Specifically, we weigh the futures-stock connectivity 
with the scaled dot-product as follows:
\begin{equation}
\text{SA}(\emb{S}^*,\emb{C}^*) = \frac{1}{\sqrt{d}} \soft\left( \emb{S}^*\emb{W}^Q_c \cdot (\emb{C}^*\emb{W}_c^K)^{\mathsf{T}} \right) \cdot \emb{C}^*\emb{W}_c^V.
\end{equation}%
$\emb{W}_c^Q$, $\emb{W}_c^K$, and $\emb{W}_c^V$ are three matrices with { $\mathbb{R}^{d \times d}$}. 
Then \texttt{CME-Layer} further concatenates multiple heads (MA) as follows:
\begin{equation}
\text{MA}(\emb{S}^*,\emb{C}^*) = \text{Concat}(\text{SA}^{\mathsf{T}}_1, \text{SA}^{\mathsf{T}}_2, \dots, \text{SA}^{\mathsf{T}}_8)\cdot \emb{W}_c^O.
\end{equation}%
In our work, the matrix $\emb{W}_c^O \in \mathbb{R}^{ 8d \times d}$ is for transformation.
Then we directly apply the Feed Forward Network (FFN) and residual connection with Layer Normalization (LN) as follows:
\begin{equation}
    \emb{S}_c^{[l+1]} = \text{LN}\Big(\emb{S}_c^{[l]} + \text{FFN}\big(\text{MA}(\emb{S}_c^{[l]}, \emb{C}^*)\big)\Big),
\end{equation}
where $\emb{S}_c^{[1]}$ is initialized by $\emb{S}^*$ and the output $\overline{\emb{S}}_c$ = $\emb{S}_c^{[8]}$. 
In our work, FFN is implemented with two linear layers and ReLU activation.
Similarly, to aggregate financial futures information, the stock can be encoded as $\overline{\emb{S}}_e = \texttt{CME-Layer}^{[8]}(\emb{S}^*, \emb{E}^*)$. 
Both $\overline{\emb{S}}_c$ and $\overline{\emb{S}}_e$ play a crucial role in capturing stock information based on their relationships with commodity and financial futures markets.
Therefore, we integrate them as the \textit{market position encoding}, which serves as the input for the subsequent \textit{Price Movement Forecasting Layer}.

\subsubsection{\textbf{Price Movement Forecasting Layer}}
The obtained embeddings $\overline{\emb{S}}_c$ and $\overline{\emb{S}}_e$ encapsulate rich futures-stock information, revealing the stocks' market positional knowledge within complex financial environments.
Therefore, the stock representations can be further updated as follows:
\begin{equation}
    \emb{S}^+ = \emb{S}^* + \overline{\emb{S}}_c + \overline{\emb{S}}_e.
\end{equation}
Based on $\emb{S}^+$, we thus further implement our second level of modules, i.e., the 8 layers of \textit{Price Movement Forecasting Layer} (\texttt{PMF-Layer}):
\begin{equation}
    \widehat{\emb{r}} = \texttt{PMF-Layer}^{[8]}(\emb{S}^+).
\end{equation}

\texttt{PMF-Layer} attentively captures the stock-stock correlations and predicts the stock price movement.
Concretely, we implement the following Multi-head Stock Attention (MSA):
\begin{equation}
\text{MSA}(\emb{S}^+) = \text{Concat}(\text{SSA}^{\mathsf{T}}_1, \text{SSA}^{\mathsf{T}}_2, \cdots, \text{SSA}^{\mathsf{T}}_8) \cdot \emb{W}_s^O,
\end{equation}
where Single-head Stock Attention (\text{SSA}) is implemented as follows:
\begin{equation}
    \text{SSA}(\emb{S}^+) = \frac{1}{\sqrt{d}}\soft\left(\emb{S}^+\emb{W}_s^Q \cdot (\emb{S}^+\emb{W}_s^K)^{\mathsf{T}} \right) \cdot \emb{S}^+\emb{W}_s^V.
\end{equation}
Here $\emb{W}_s^Q$, $\emb{W}_s^K$, $\emb{W}_s^V$ are transformation matrices with { $\mathbb{R}^{d \times d}$}, and $\emb{W}_s^O \in \mathbb{R}^{ 8d \times d}$ .
We then follow the standard forward procedure~\cite{vaswani2017attention} with the FFN and LN as follows:
\begin{equation}
    {\emb{S}^+}^{[l+1]} = \text{LN}\big(\widetilde{\emb{S}^+}^{[l]} + \text{FFN}(\widetilde{\emb{S}^+}^{[l]})\big), 
\end{equation}
where $\widetilde{\emb{S}^+}^{[l]}$ is obtained via:
\begin{equation}
\widetilde{\emb{S}^+}^{[l]} = \text{LN}\big({\emb{S}^+}^{[{l}]} + \text{MSA}({\emb{S}^+}^{[{l}]})\big).
\end{equation}
Finally, the movements $\widehat{\emb{r}}$ are derived by adopting the linear transformation to the output ${\emb{S}^+}^{[8]}$.
In summary, our Bi-level Dense Pricing Transformer captures both the futures-to-stock and stock-wise correlations, enabling cross-market synergistic approach to stock price forecasting.

\subsection{Pseudo-volatility Optimization}
\label{sec:volatility}
Due to varying positions in the stock market, different stocks exhibit differing levels of revenue-generating capability and resilience to market risks.
These factors result in their distinct levels of stock price volatility.
Stock volatility indicates the stability of stock price fluctuations and uncertainty in predictions.
Intuitively, higher volatility often signifies greater uncertainty in price forecasting.
Therefore, in this work, we are motivated to capture this concept and propose learning the ``\textit{pseudo-volatility}'' for more accurate stock price forecasting. 

\subsubsection{\textbf{Pseudo-volatility Estimation.}}
To capture such volatility, one possible solution is to leverage Bayesian deep learning~\cite{wilson2020bayesian,maddox2019simple}, which originally offers a practical framework for modeling uncertainty~\cite{kendall2017uncertainties}.
Inspired by these works, we propose estimating the stock pseudo-volatility within the Bayesian framework to have the following deep neural architectures, namely \texttt{PV-Estimator}:
\begin{equation}
 \emb{\gamma} = \texttt{PV-Estimator}(\emb{S}), \text{ where } \emb{\gamma} \in \mathbb{R}^{k_s \times d}.
\end{equation}
$S$ is the raw stock market data and $\emb{\gamma}$ is the estimated pseudo-volatility.
Our designed \texttt{PV-Estimator} differs from regular deterministic neural networks by incorporating volatility modeling and their variational inference. 
Specifically, we first process $\emb{S}$ via a two-layer MLP with ReLU activation, denoted by \texttt{$\texttt{MLP}^{[2]}$}, as follows:
\begin{equation}
\label{eq:mlp}
\emb{V} = \texttt{MLP}^{[2]}(\emb{S}), \text{ where } \emb{V} \in \mathbb{R}^{ t \times k_s \times d}.
\end{equation}
Then we pass it through the eight-layer vanilla Transformer~\cite{vaswani2017attention}, denoted by \texttt{Trm-layer$^{[8]}$}:
\begin{equation}
\label{eq:trm}
\widetilde{\emb{V}} = \texttt{Trm-Layer}^{[8]}(\emb{V}), \text{ where } \widetilde{\emb{V}} \in \mathbb{R}^{t \times k_s \times d}.
\end{equation}
Then the pseudo-volatility $\emb{\gamma}$ is empirically estimated with the mean-pooling operation as:
\begin{equation}
    \emb{\gamma} = \texttt{Mean-Pooling}( \widetilde{\emb{V}}), \text{ where } \emb{\gamma} \in \mathbb{R}^{k_s \times d}.
\end{equation}

In our work, we utilize Monte Carlo dropout throughout Eqn.'s~(\ref{eq:mlp}) and (\ref{eq:trm}) to achieve the variational volatility estimation.
Intuitively, $\emb{\gamma}$ is calculated via an independent computational pipeline mainly to capture the hidden volatility information from raw market data in a less biased manner. 
Since stock assets exhibit varying levels of volatility, and given that a higher value of $\emb{\gamma}$ indicates a higher uncertainty, the model should adapt its learning process accordingly. 
Along with the forecasted price discussed in the previous section, we propose a {volatility-aware learning objective} to eventually enable a more fine-grained optimization approach as follows.

\subsubsection{\textbf{Volatility-aware Regression Optimization}}
In conventional learning paradigms, regression objectives like mean squared error (MSE) are typically used.
However, since we consider pseudo-volatility to distinguish stocks with varying prediction confidences, we incorporate this knowledge into model optimization.
Intuitively, higher pseudo-volatility $\emb{\gamma}$ indicates lower confidence in price forecasting and should therefore reduce its contribution to the accumulated loss.
A straightforward way to achieve this is by:
\begin{equation}
\label{eq:hete_mse}
 \mathcal{L} = \frac{1}{k_s}\sum \frac{(\emb{r} - \widehat{\emb{r}})^2}{{\emb{\gamma}}},
\end{equation}
where $\emb{r}$ is the ground-truth prices of all $k_s$ stocks at the ($t+1$)-th time step.
Eqn.~(\ref{eq:hete_mse}) differentiates the standard MSE where the MSE essentially assumes equal variance for all stock samples.

However, Eqn.~(\ref{eq:hete_mse}) still have some inadequacies. Firstly, during the loss minimization of Eqn.~(\ref{eq:hete_mse}), it may easily minimize the loss by optimizing $\emb{\gamma}$ into negative values, which is inappropriate for $\emb{\gamma}$ as it should be a positive value.
To fix this issue, we simply apply the numerical scaling with exponentiation as:
\begin{equation}
\label{eq:better1}
     \mathcal{L} = \frac{1}{k_s}\sum \frac{(\emb{r} - \widehat{\emb{r}})^2}{{\exp(\emb{\gamma})}}.
\end{equation}
Secondly, we want to reward the stock samples with low pseudo-volatility. 
However, our design may not perfectly align with Eqn.~(\ref{eq:better1}) as it may also maximize $\emb{\gamma}$ values, rather than solely optimizing the discrepancy between $\emb{r}$ and $\widehat{\emb{r}}$, which may disturb the optimization direction.
Therefore, we update it with the following regularization:
\begin{equation}
\label{eq:regular}
\mathcal{L} = \frac{1}{k_s}\sum \frac{(\emb{r} - \widehat{\emb{r}})^2}{{\exp({\emb{\gamma}})}} + \emb{\gamma}.
\end{equation}
Lastly, due to the stochastic process of variational volatility estimation in \texttt{PV-Estimator}, we further propose using ensembling to stabilize the pseudo-volatility estimation and optimization:
\begin{equation}
\label{eq:final}
\mathcal{L} = \frac{1}{k_sH} \sum^H_{h=1} \sum \Big(\frac{(\emb{r} - \widehat{
\emb{r}})^2}{{\exp(\emb{\gamma}_h)}} + \emb{\gamma}_h\Big).
\end{equation}
$\emb{\gamma}_h$ is output from an independent \texttt{PV-Estimator}.
In our work, setting $H=2$ already achieves satisfactory performance.
As our empirical analysis in~\cref{exp:loss} demonstrates, compared to our initial design in Eqn.~(\ref{eq:hete_mse}), Eqn.~(\ref{eq:final}) eventually provides more appropriate loss scaling, balanced optimization directions, and improved movement forecasting performance.

\section{Experiments}
\label{sec:exp}
We aim to answer the following research questions:
\begin{itemize}[leftmargin=*]
\item \textbf{RQ1}: How does our model help in real-world trading scenarios to enhance proprietary profitability?
\item \textbf{RQ2}: How does our model perform compared to existing models on real-world datasets?
\item \textbf{RQ3}: How to systematically evaluate designs within \model?
\end{itemize}

\subsection{Proprietary Backtesting Evaluation (RQ1)}
\subsubsection{\textbf{Overview.}}
To assess how our model supports real-world trading, we introduce two evaluation approaches:
\begin{itemize}[leftmargin=*]
\item \textbf{Task 1: trading executor}. We directly follow our model's predicted price movement as signals to trigger stock trades, and then integrate these trades into our internal platform for backtesting.

\item \textbf{Task 2: alpha factor producer.} The other approach incorporates the model's output features as additional alpha factors to enhance our holistic trading strategies. We then evaluate whether these factors can contribute to improved system performance in the general portfolio investment.
\end{itemize}

\begin{table}[t]
\centering
\caption{Data statistics for proprietary trading evaluation. ``\#'' denotes the size. ``C-futures'' and ``F-futures'' denote commodity and financial futures.
We use quarter time for data splitting, e.g., ``08Q1'' means the first quarter of 2008.
``Ins.'' and ``Trans.'' denote the ``Instruments'' and ``Transaction days''. }
\label{tb:internalData}
\resizebox{\linewidth}{!}{
\begin{tabular}{c|cccc|cccc}
\specialrule{.1em}{0em}{0em}
 \multirow{4}{*}{\diagbox[innerwidth=1.2cm]{\normalsize Asset}{\normalsize {\quad Time}}}
& \multicolumn{4}{c|}{Task 1} & \multicolumn{4}{c}{Task 2} \\ 
~ &\multicolumn{2}{c}{Training} &\multicolumn{2}{c|}{Evaluation} &\multicolumn{2}{c}{Training} &\multicolumn{2}{c}{Evaluation} \\
~ & \multicolumn{2}{c}{08Q1-16Q4} & \multicolumn{2}{c|}{17Q1-20Q3} & \multicolumn{2}{c}{17Q1-19Q2} & \multicolumn{2}{c}{19Q3-20Q3} \\
~ & \#Ins. & \#Trans. & 
\#Ins. & \#Trans. & 
\#Ins. & \#Trans. & 
\#Ins. & \#Trans.\\
\hline 
\textbf{C-futures}     & 2,770   & 2,190 & 2,243 & 871 & 1,588& 606& 1,149& 266\\
\textbf{F-futures}     & 424     & 2,190 & 344 & 871 & 255& 606& 142& 266\\
\cline{1-9}
\textbf{CSI300}     & 300   & 2,190  & 300 & 871 & 300& 606& 300& 266\\
\textbf{CSI500}     & 500   & 2,190 & 500 & 871 & 500& 606& 500& 266\\
\textbf{CSI1000}     & 1,000  & 2,190 & 1,000 & 871 & 1,000& 606& 1,000& 266\\
\specialrule{.1em}{0em}{0em}
\end{tabular}}
\end{table}

\subsubsection{\textbf{Evaluation Data.}}
We use real-world financial market data, i.e., futures and stocks, for model training and evaluation.
We collect the daily frequency futures data, i.e., commodity and financial futures, from the Chinese futures market.
For stock market data, we include two most frequently used index data, i.e., CSI300 and CSI500, and one most diverse index, i.e., CSI1000, from Shanghai Stock Exchange and the Shenzhen Stock Exchange.
CSI300 and CSI500 are capitalization-weighted stock market indexes designed to replicate the performance of the top traded 300 and 500 stocks.
And CSI1000 focuses on small-cap companies.
We collect all data within 2008-2020 and use the data split of 2008-2017 for training and 2018-2020 for evaluation. 
Data statistics are reported in Table~\ref{tb:internalData}.

\subsubsection{\textbf{Evaluation Metrics.}}
We introduce the following series of evaluation metrics: 
(1) Annualized Return (AR), (2) Winning Rate (WR), (3) Sortino Ratio (SoR), (4) Sharpe Ratio (ShR), (5) Maximum Drawdown (MD), (6) Maximum Drawdown Duration (MD-D), (7) Turnover Rate (TR). 
\textit{Note that for the first four metrics, a higher value indicates better performance, whereas for the last three metrics, a lower value suggests a more favorable outcome.}
Due to page limit, we explain them in detail in Appendix~\ref{app:metrics} with evaluation configuration details in~\cref{app:setup}.

\begin{table}[t]
\tabcolsep=0.1cm
\centering
\vspace{-0.3em}
\caption{Market backtesting of stock trading executor. (1) The strategy refers to whether it is for ``Hedged Return'' or ``Spot Return''. (2) The symbols $\uparrow$ and $\downarrow$ denote cases where higher and lower values, respectively, indicate better performance. (3) Colors indicate better-performing cases, i.e., \colorbox{high1}{better}.}
\label{tb:hedged_csi}
\resizebox{\linewidth}{!}{
\begin{tabular}{c|cccccccc}
\specialrule{.1em}{0em}{0em}
Data & Strategy & AR $\uparrow$ & WR $\uparrow$ & SoR $\uparrow$ & ShR $\uparrow$ & MD $\downarrow$ & MD-D $\downarrow$ & TR $\downarrow$  \\
\hline  
\multirow{2}{*}{CSI300} & Hedged & 0.0336 & 0.5071 & 0.5401 & 0.3205 & \fst{0.2544} & \fst{196} &  \multirow{2}{*}{0.1057}  \\
& Spot & \fst{0.1449} & \fst{0.5259} & \fst{1.1547} & \fst{0.7002} & 0.3617 & 271 & ~ \\
\cline{1-9}
\multirow{2}{*}{CSI500} & Hedged & 0.1602 & 0.5112 & \fst{3.1518} & \fst{1.6386} & \fst{0.2783} & 224 & \multirow{2}{*}{0.0357} \\
& Spot & \fst{0.1718} & \fst{0.5206} & 1.4232 & 0.8412 & 0.3366 & \fst{200} & ~ \\
\cline{1-9}
\multirow{2}{*}{CSI1000} & Hedged & \fst{0.3261} & \fst{0.5512} & \fst{4.4248} & \fst{2.3853} & \fst{0.1355} & 118 & \multirow{2}{*}{0.0429}  \\
& Spot & 0.2837 & 0.5312 & 1.8849 & 1.1036 & 0.2930 & \fst{108} & ~  \\
\specialrule{.1em}{0em}{0em}
\end{tabular}
}
\end{table}

\begin{figure}[t]
    \vspace{-0.4cm}
    \hspace{-1em}
    \begin{minipage}{0.24\textwidth}
        \includegraphics[width=\textwidth]{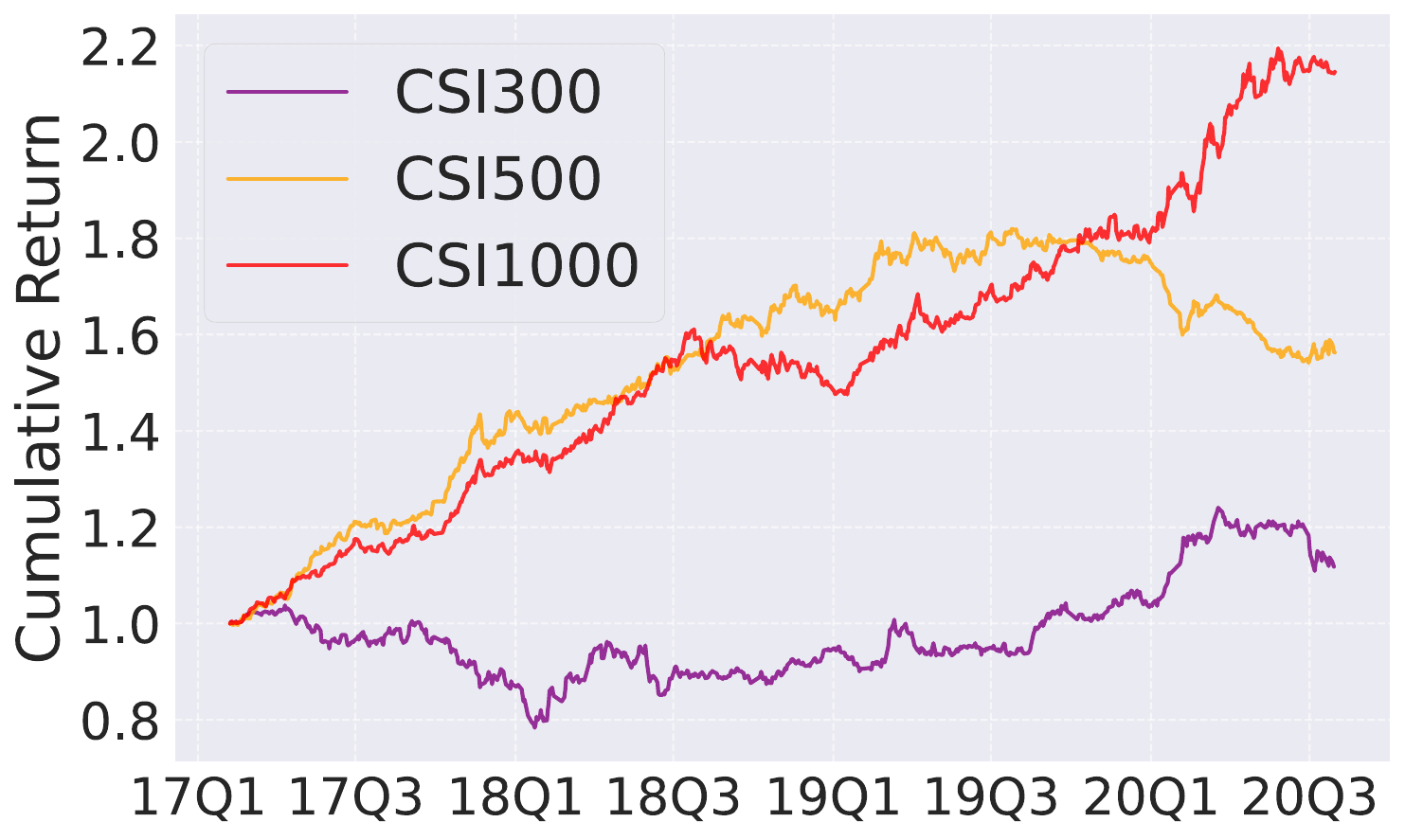}
        \vspace{-1.5em}
        \subcaption{Hedged return curve.}
    \end{minipage}    
    \begin{minipage}{0.24\textwidth}
        \includegraphics[width=\textwidth]{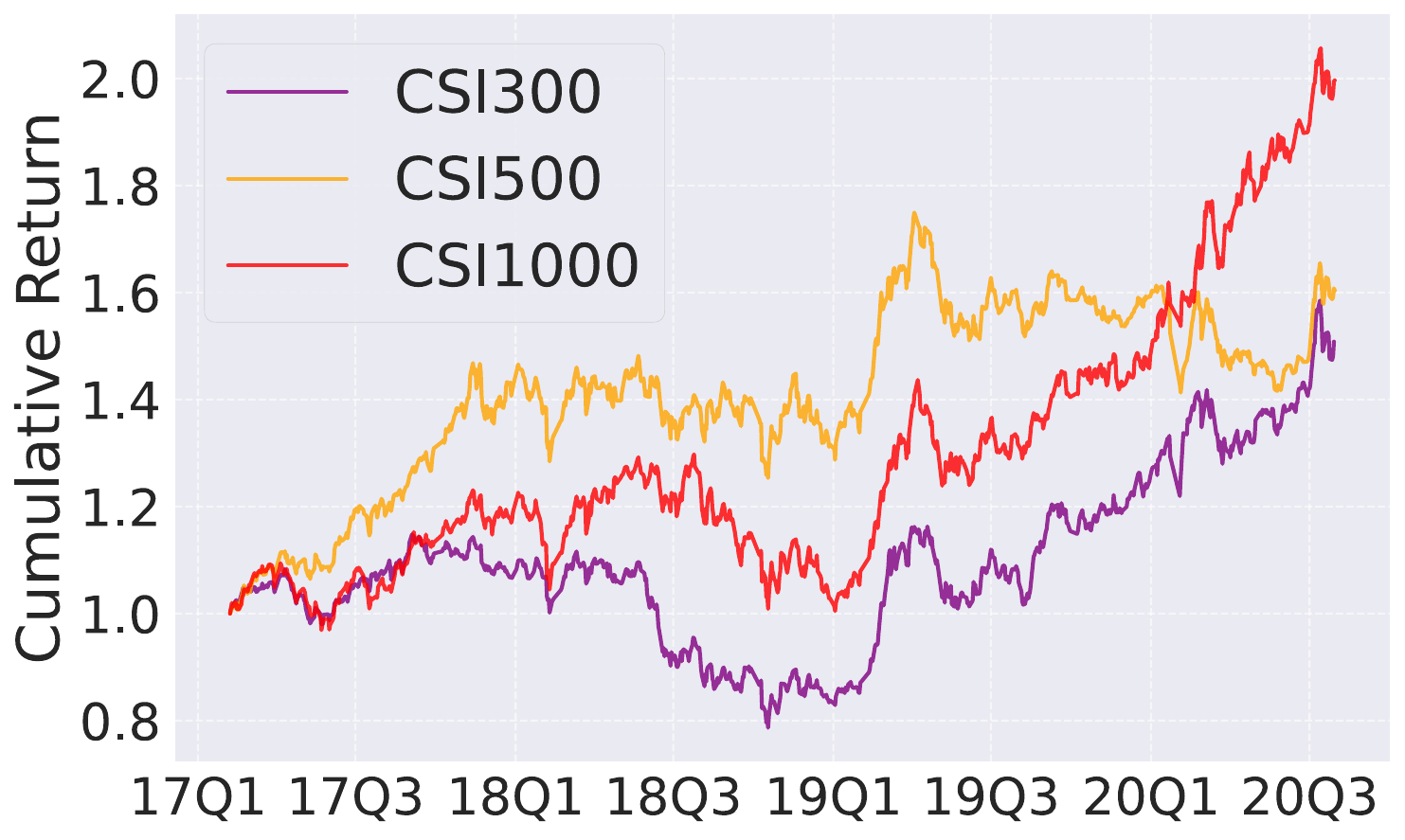}
        \vspace{-1.5em}
        \subcaption{Spot return curve.}
    \end{minipage}
    \caption{Portfolio yield curves of our trading executor.}
    \label{fig:single_curve}
\end{figure}

\begin{figure*}[t]
    \hspace{-1em}
    \begin{minipage}{0.33\textwidth}
        \includegraphics[width=\textwidth]{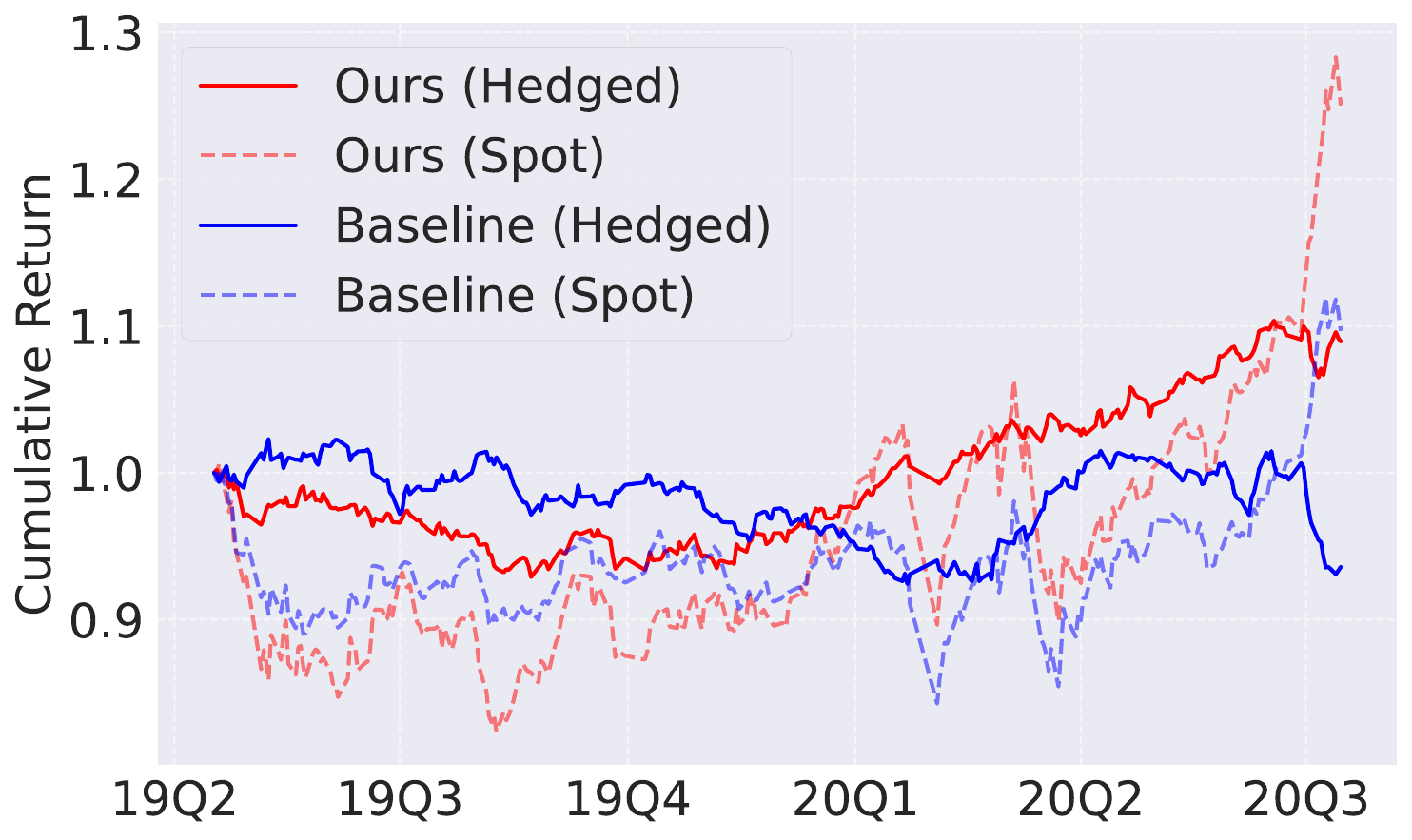}
        \vspace{-1.5em}
        \subcaption{CSI300 yield curve.}
        \label{fig:CSI300_1}
    \end{minipage}%
    \begin{minipage}{0.33\textwidth}
        \includegraphics[width=\textwidth]{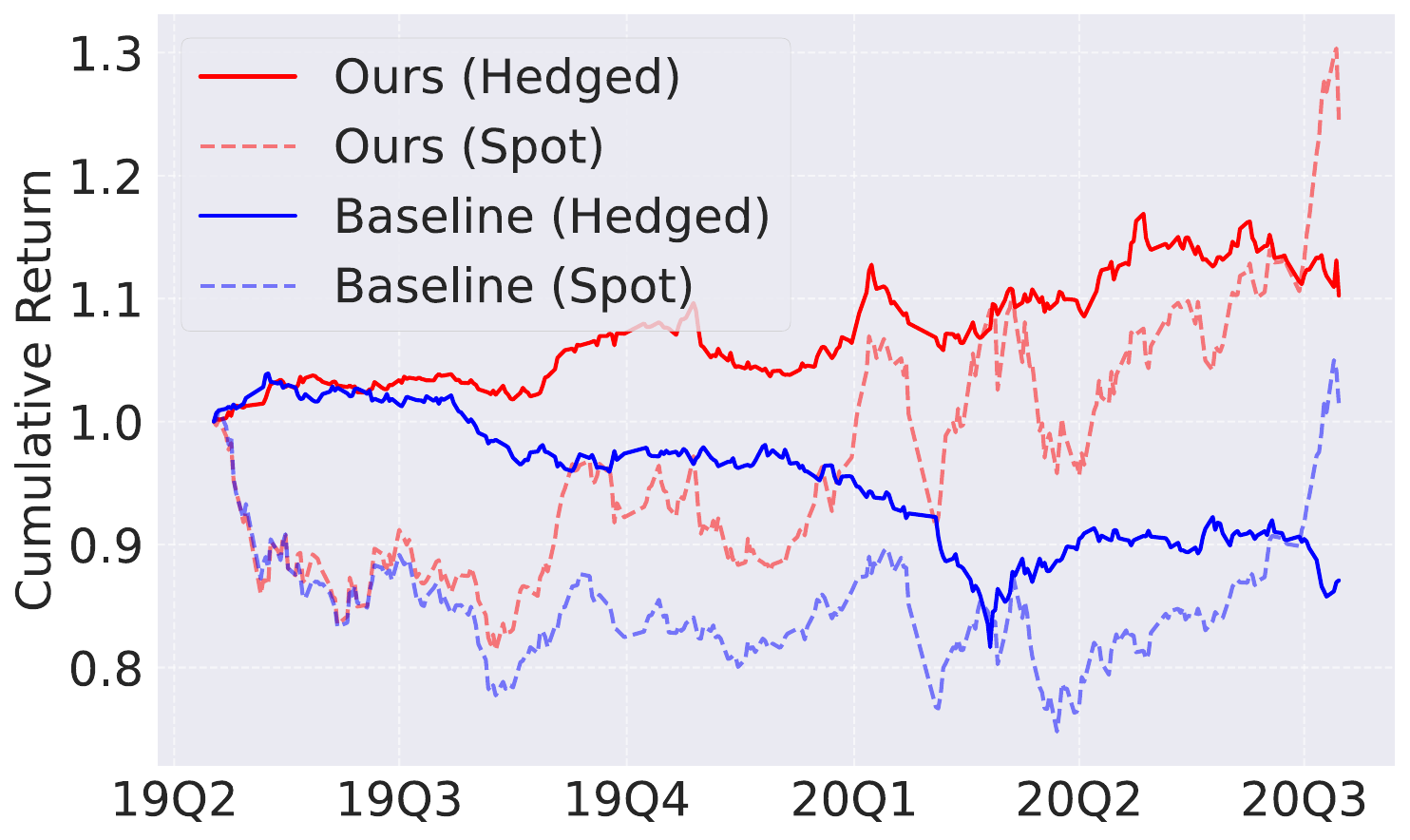}
        \vspace{-1.5em}
        \subcaption{CSI500 yield curve.}
        \label{fig:CSI500_1}
    \end{minipage}%
    \begin{minipage}{0.33\textwidth}
        \includegraphics[width=\textwidth]{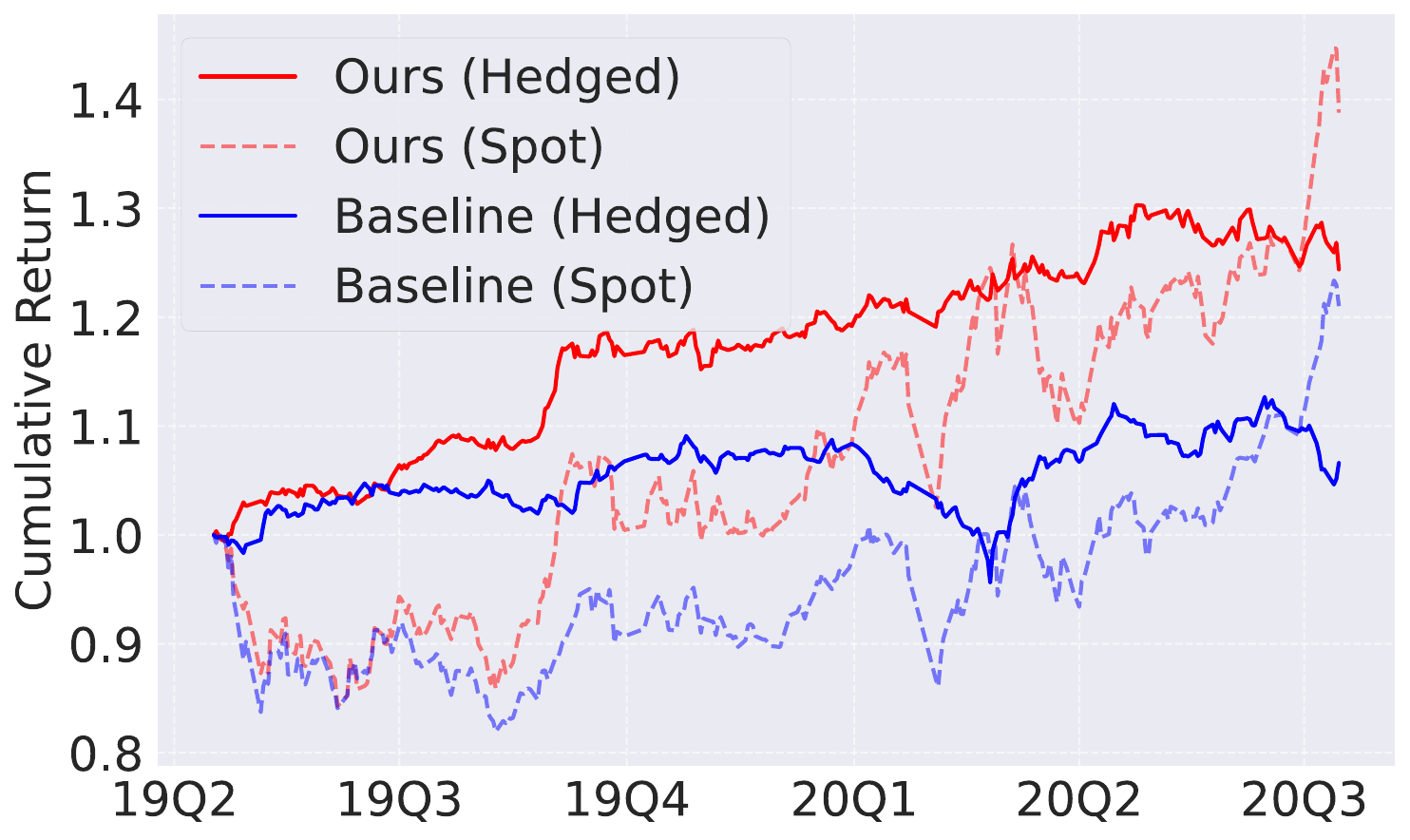}
        \vspace{-1.5em}
        \subcaption{CSI1000 yield curve.}
        \label{fig:CSI1000_1}
    \end{minipage}
    \caption{Yield curves of portfolio investments with (in warm colors) or without (in cold colors) our additional Alpha factors.}
    \label{fig:CSI_comparison}
    \vspace{-0.5cm}
\end{figure*}

\subsubsection{\textbf{Task 1: Stock Trading Executor}}
As we briefly introduced, we first rely on our model's price predictions to execute stock trades, by following our proprietary protocol: \textit{enhanced indexing with daily rebalancing}.
This is widely adopted in quantitative active management due to its capacity to manage large amounts of capital and relatively stable returns.
Specifically, we rank all candidate stocks based on our model’s predictions, select a certain number of stocks to buy under a specific budget, and adjust the portfolio daily based on the model's new predictions (either increasing or decreasing positions). 
We apply two types of trading strategies for \textit{hedged return} and \textit{spot return}. 
To achieve stable returns, we could neutralize the market risks by simultaneously \textit{shorting index derivatives}. 
Thus, our hedged return depends on the relative performance between the selected stocks and the index shorts.
Conversely, the spot return applies without hedging.
We curve the whole portfolio backtesting yields from 2017Q1 to 2020Q3 in Figure~\ref{fig:single_curve} and report metrics in Table~\ref{tb:hedged_csi}.
From these detailed metrics, we notice that:
\begin{itemize}[leftmargin=*]
    \item For CSI300 and CSI500 indices, we notice that the spot return strategy yields better performance with higher annual returns and winning rates.
    Since these two indices comprise more ``blue chip'' stocks with lower volatility, the spot return strategy is generally a suitable choice for achieving higher profits.
    
    \item Compared to CSI300 and CSI500, the CSI1000 index includes a higher proportion of ``small cap stocks'', which are associated with larger price volatility. 
    Therefore, the hedged return strategy demonstrates better performance in backtesting and may be more beneficial in practice for hedging market risks.
\end{itemize}

\begin{table}[t]
\tabcolsep=0.05cm
\centering
\vspace{-0.5em}
\caption{Evaluation of Alpha factor production. (1) ``$\checkmark$'' and ``$\times$'' respectively denote the cases where the additional factors produced from our model are integrated or not.}
\label{tb:alpha}
\resizebox{\linewidth}{!}{
\begin{tabular}{c|ccccccccc}
\specialrule{.1em}{0em}{0em}
    Data & Strategy & Alpha & AR $\uparrow$ & WR $\uparrow$ & SoR $\uparrow$ & ShR $\uparrow$ & MD $\downarrow$ & MD-D $\downarrow$ & TR $\downarrow$ \\
\hline  
\multirow{4}{*}{CSI300} & \multirow{2}{*}{Hedged} & $\times$ & -0.0891 & 0.4536 & -1.8021 & -1.0611 & 0.1365 & 289 & 0.0398 \\
    & & $\checkmark$ & \fst{0.0715} & \fst{0.5298} & \fst{1.5488} & \fst{0.9280} & \fst{0.0731} & \fst{87} & \fst{0.0168} \\
\cline{2-10}
~ & \multirow{2}{*}{Spot} & $\times$ & 0.0393 & 0.5397 & 0.4775 & 0.3175 & 0.1427 & 226 & 0.0398 \\
    & & $\checkmark$ & \fst{0.1999} & \fst{0.5497} & \fst{1.3434} & \fst{0.8929} & \fst{0.1826} & \fst{76} & \fst{0.0168} \\
\hline  
\multirow{4}{*}{CSI500} & \multirow{2}{*}{Hedged} & $\times$ & -0.1114 & 0.4437 & -1.9357 & -1.1405 & 0.2085 & 194 & 0.0515 \\
    & & $\checkmark$ & \fst{0.0818} & \fst{0.5132} & \fst{1.5821} & \fst{0.9276} & \fst{0.0696} & \fst{15} & \fst{0.0190} \\
\cline{2-10}
~ & \multirow{2}{*}{Spot} & $\times$ & 0.0029 & 0.5001 & 0.1843 & 0.1240 & 0.2274 & 224 & 0.0515 \\
    & & $\checkmark$ & \fst{0.1961} & \fst{0.5497} & \fst{1.2341} & \fst{0.7940} & \fst{0.1880} & \fst{76} & \fst{0.0190} \\
\hline  
\multirow{4}{*}{CSI1000} & \multirow{2}{*}{Hedged} & $\times$ & 0.0528 & 0.4868 & 1.1124 & 0.6058 & 0.1348 & 80 & 0.0937  \\
    & & $\checkmark$ & \fst{0.1947} & \fst{0.5232} & \fst{3.5369} & \fst{1.9198} & \fst{0.0591} & \fst{53} & \fst{0.0225} \\
\cline{2-10}
~ & \multirow{2}{*}{Spot} & $\times$ & 0.1678 & 0.5262 & 1.1034 & 0.7555 & 0.1824 & 76 & 0.0937 \\
    & & $\checkmark$ & \fst{0.3097} & \fst{0.5397} & \fst{1.8087} & \fst{1.1243} & \fst{0.1649} & \fst{12} & \fst{0.0225} \\
\specialrule{.1em}{0em}{0em}
\end{tabular}}
\end{table}

\subsubsection{\textbf{Task 2: Alpha Factor Producer}}
In quantitative finance, analysts normally use a combination of trading signals, i.e., \textit{Alpha factors}, to manage risk more effectively and make data-driven portfolio investment decisions.
To further assess our \model, we specifically validate its capability in producing Alpha factors.
We integrate our model into our internal quantitative analysis framework, serving as a source of factor production.
The output factors are then utilized as inputs in the downstream portfolio investment system.
Based on the existing factors, we then compare the performances between solely using these existing factors and using our new factors in addition.
The backtesting yield curves on three stock pools are shown in Figure~\ref{fig:CSI_comparison} and detailed evaluation results are reported in Table~\ref{tb:alpha}.
We have the following twofold observations:
\begin{itemize}[leftmargin=*]
\item We observe from Figure~\ref{fig:CSI_comparison} that, the warm-colored curves (for both hedged and spot return strategies), representing the cases with our model's Alpha factors, exhibit a more stable upward trend over time compared to our original baseline, i.e., in cold-colored lines. 
This suggests that our model consistently provides positive returns with less volatility, making it a reliable choice for achieving steady investment growth in these stock pools.

\item As shown in Table~\ref{tb:alpha}, strategies for both hedged and spot returns that incorporate our additional alpha factors consistently outperform the original settings across all evaluation metrics.  
Furthermore, the hedged return strategy is usually more favorable in practice due to its anticipated stability.
Additionally, our original factor settings on CSI300 and CSI500 may yield unsatisfied performances, e.g., negative annual returns;
in contrast, our model effectively turns losses into profits, 
which demonstrates its capability to generate impactful and effective Alpha factors.
\end{itemize}

\begin{table*}[t]
\renewcommand\arraystretch{1.1}
\caption{Performance comparison across different models on CSI100, CSI300, and CSI500 data. Colors indicate the best and second-best performing models, i.e., \colorbox{high1}{best} and \colorbox{secondhigh}{second-best}.}
\label{tab:public_performance}
\tabcolsep=0.05cm
\centering
\small
\resizebox{1\textwidth}{!}{
\begin{tabular}{c|cccc|cccc|cccc}
\specialrule{.1em}{0em}{0em}
\multirow{2}{*}{\textbf{Model}} & \multicolumn{4}{c|}{\textbf{CSI100}} & \multicolumn{4}{c|}{\textbf{CSI300}} & \multicolumn{4}{c}{\textbf{CSI500}} \\
\cline{2-13}
& \textbf{IC $\uparrow$} & \textbf{IR$_{\text{IC}}$ $\uparrow$} & \textbf{RIC $\uparrow$} & \textbf{IR$_{\text{Rank IC}}$ $\uparrow$} & \textbf{IC $\uparrow$} & \textbf{IR$_{\text{IC}}$ $\uparrow$} & \textbf{RIC $\uparrow$} & \textbf{IR$_{\text{Rank IC}}$ $\uparrow$} & \textbf{IC $\uparrow$} & \textbf{IR$_{\text{IC}}$ $\uparrow$} & \textbf{RIC $\uparrow$} & \textbf{IR$_{\text{Rank IC}}$ $\uparrow$}  \\
\hline
LSTM & 0.0280 {\footnotesize $\pm$ 0.00} & 0.1489 {\footnotesize $\pm$ 0.02} & 0.0401 {\footnotesize $\pm$ 0.00} & 0.2207 {\footnotesize $\pm$ 0.02} & 0.0323 {\footnotesize $\pm$ 0.00} & 0.2296 {\footnotesize $\pm$ 0.04} & 0.0411 {\footnotesize $\pm$ 0.00} & 0.3401 {\footnotesize $\pm$ 0.03} & 0.0389 {\footnotesize $\pm$ 0.00} & 0.3904 {\footnotesize $\pm$ 0.05} & 0.0493 {\footnotesize $\pm$ 0.00} & 0.5310 {\footnotesize $\pm$ 0.03} \\
GRU & 0.0299 {\footnotesize $\pm$ 0.00} & 0.1667 {\footnotesize $\pm$ 0.02} & 0.0401 {\footnotesize $\pm$ 0.00} & 0.2284 {\footnotesize $\pm$ 0.02} & 0.0329 {\footnotesize $\pm$ 0.00} & 0.2403 {\footnotesize $\pm$ 0.05} & 0.0423 {\footnotesize $\pm$ 0.00} & 0.3399 {\footnotesize $\pm$ 0.03} & 0.0414 {\footnotesize $\pm$ 0.00} & 0.3919 {\footnotesize $\pm$ 0.04} & \snd{0.0565} {\footnotesize $\pm$ 0.00} & 0.5812 {\footnotesize $\pm$ 0.03} \\
Transformer & 0.0239 {\footnotesize $\pm$ 0.01} & 0.1411 {\footnotesize $\pm$ 0.01} & 0.0349 {\footnotesize $\pm$ 0.00} & 0.2124 {\footnotesize $\pm$ 0.03} & 0.0254 {\footnotesize $\pm$ 0.00} & 0.2040 {\footnotesize $\pm$ 0.02} & 0.0427 {\footnotesize $\pm$ 0.00} & 0.3181 {\footnotesize $\pm$ 0.02} & 0.0302 {\footnotesize $\pm$ 0.00} & 0.2884 {\footnotesize $\pm$ 0.03} & 0.0472 {\footnotesize $\pm$ 0.00} & 0.4811 {\footnotesize $\pm$ 0.03} \\
ALSTM & 0.0364 {\footnotesize $\pm$ 0.00} & 0.2124 {\footnotesize $\pm$ 0.03} & 0.0423 {\footnotesize $\pm$ 0.01} & 0.2562 {\footnotesize $\pm$ 0.02} & 0.0371 {\footnotesize $\pm$ 0.01} & 0.2697 {\footnotesize $\pm$ 0.05} & 0.0455 {\footnotesize $\pm$ 0.01} & 0.3786 {\footnotesize $\pm$ 0.06} & 0.0396 {\footnotesize $\pm$ 0.01} & 0.3946 {\footnotesize $\pm$ 0.05} & 0.0562 {\footnotesize $\pm$ 0.00} & 0.5576 {\footnotesize $\pm$ 0.04} \\
SFM & 0.0344 {\footnotesize $\pm$ 0.01} & 0.1776 {\footnotesize $\pm$ 0.03} & 0.0436 {\footnotesize $\pm$ 0.01} & 0.2349 {\footnotesize $\pm$ 0.02} & 0.0370 {\footnotesize $\pm$ 0.00} & 0.2879 {\footnotesize $\pm$ 0.04} & 0.0463 {\footnotesize $\pm$ 0.00} & 0.3775 {\footnotesize $\pm$ 0.04} & 0.0353 {\footnotesize $\pm$ 0.00} & 0.3007 {\footnotesize $\pm$ 0.04} & 0.0510 {\footnotesize $\pm$ 0.00} & 0.4728 {\footnotesize $\pm$ 0.03} \\
TCN & 0.0179 {\footnotesize $\pm$ 0.00} & 0.1132 {\footnotesize $\pm$ 0.02} & 0.0146 {\footnotesize $\pm$ 0.00} & 0.0813 {\footnotesize $\pm$ 0.02} & 0.0279 {\footnotesize $\pm$ 0.00} & 0.2181 {\footnotesize $\pm$ 0.01} & 0.0421 {\footnotesize $\pm$ 0.04} & 0.3429 {\footnotesize $\pm$ 0.01} & 0.0103 {\footnotesize $\pm$ 0.02} & 0.0933 {\footnotesize $\pm$ 0.08} & 0.0087 {\footnotesize $\pm$ 0.01} & 0.0870 {\footnotesize $\pm$ 0.07} \\
TabNet & 0.0279 {\footnotesize $\pm$ 0.00} & 0.1596 {\footnotesize $\pm$ 0.01} & 0.0360 {\footnotesize $\pm$ 0.00} & 0.2142 {\footnotesize $\pm$ 0.02} & 0.0199 {\footnotesize $\pm$ 0.01} & 0.1477 {\footnotesize $\pm$ 0.07} & 0.0351 {\footnotesize $\pm$ 0.00} & 0.2693 {\footnotesize $\pm$ 0.05} & 0.0321 {\footnotesize $\pm$ 0.00} & 0.3562 {\footnotesize $\pm$ 0.03} & 0.0406 {\footnotesize $\pm$ 0.00} & 0.4425 {\footnotesize $\pm$ 0.03} \\
TFT & 0.0278 {\footnotesize $\pm$ 0.01} & 0.1333 {\footnotesize $\pm$ 0.03} & 0.0114 {\footnotesize $\pm$ 0.02} & 0.0551 {\footnotesize $\pm$ 0.01} & 0.0338 {\footnotesize $\pm$ 0.00} & 0.1999 {\footnotesize $\pm$ 0.03} & 0.0129 {\footnotesize $\pm$ 0.01} & 0.0913 {\footnotesize $\pm$ 0.04} & 0.0398 {\footnotesize $\pm$ 0.01} & 0.2900 {\footnotesize $\pm$ 0.04} & 0.0117 {\footnotesize $\pm$ 0.01} & 0.0894 {\footnotesize $\pm$ 0.09} \\
Localformer & 0.0289 {\footnotesize $\pm$ 0.00} & 0.1747 {\footnotesize $\pm$ 0.02} & 0.0351 {\footnotesize $\pm$ 0.00} & 0.2147 {\footnotesize $\pm$ 0.01} & 0.0373 {\footnotesize $\pm$ 0.00} & 0.2983 {\footnotesize $\pm$ 0.03} & 0.0488 {\footnotesize $\pm$ 0.00} & 0.3869 {\footnotesize $\pm$ 0.03} & 0.0362 {\footnotesize $\pm$ 0.00} & 0.3374 {\footnotesize $\pm$ 0.03} & 0.0551 {\footnotesize $\pm$ 0.00} & 0.5584 {\footnotesize $\pm$ 0.03} \\
TRA & 0.0458 {\footnotesize $\pm$ 0.01} & 0.2543 {\footnotesize $\pm$ 0.01} & \snd{0.0534} {\footnotesize $\pm$ 0.01} & \snd{0.3037} {\footnotesize $\pm$ 0.03} & 0.0445 {\footnotesize $\pm$ 0.01} & 0.3653 {\footnotesize $\pm$ 0.05} & \snd{0.0533} {\footnotesize $\pm$ 0.00} & \snd{0.4403} {\footnotesize $\pm$ 0.03} & 0.0396 {\footnotesize $\pm$ 0.00} & \snd{0.4133} {\footnotesize $\pm$ 0.03} & 0.0529 {\footnotesize $\pm$ 0.01} & \snd{0.5849} {\footnotesize $\pm$ 0.05} \\
\hline
XGBoost & \snd{0.0548} {\footnotesize $\pm$ 0.01} & 0.2913 {\footnotesize $\pm$ 0.03} & 0.0473 {\footnotesize $\pm$ 0.00} & 0.2977{\footnotesize $\pm$ 0.02} & 0.0500 {\footnotesize $\pm$ 0.00} & 0.3767 {\footnotesize $\pm$ 0.00} & 0.0511 {\footnotesize $\pm$ 0.00} & 0.4344 {\footnotesize $\pm$ 0.00} & 0.0409 {\footnotesize $\pm$ 0.00} & 0.3428 {\footnotesize $\pm$ 0.01} & 0.0428 {\footnotesize $\pm$ 0.00} & 0.4071 {\footnotesize $\pm$ 0.01} \\
CatBoost & 0.0552 {\footnotesize $\pm$ 0.00} & 0.2811 {\footnotesize $\pm$ 0.01} & 0.0455 {\footnotesize $\pm$ 0.00} & 0.2639 {\footnotesize $\pm$ 0.01} & 0.0494 {\footnotesize $\pm$ 0.00} & 0.3467 {\footnotesize $\pm$ 0.00} & 0.0473 {\footnotesize $\pm$ 0.00} & 0.3507 {\footnotesize $\pm$ 0.01} & \snd{0.0419} {\footnotesize $\pm$ 0.00} & 0.3324 {\footnotesize $\pm$ 0.01} & 0.0423 {\footnotesize $\pm$ 0.00} & 0.3770 {\footnotesize $\pm$ 0.02} \\
LightGBM & 0.0403 {\footnotesize $\pm$ 0.01} & 0.2391 {\footnotesize $\pm$ 0.04} & 0.0409 {\footnotesize $\pm$ 0.00} & 0.2515 {\footnotesize $\pm$ 0.04} & 0.0466 {\footnotesize $\pm$ 0.00} & 0.3790 {\footnotesize $\pm$ 0.01} & 0.0510 {\footnotesize $\pm$ 0.01} & 0.4037 {\footnotesize $\pm$ 0.01} & 0.0381 {\footnotesize $\pm$ 0.00} & 0.3654 {\footnotesize $\pm$ 0.03} & 0.0482 {\footnotesize $\pm$ 0.00} & 0.4910 {\footnotesize $\pm$ 0.02} \\
DoubleEnsemble & 0.0478 {\footnotesize $\pm$ 0.00} & \snd{0.2933} {\footnotesize $\pm$ 0.00} & 0.0461 {\footnotesize $\pm$ 0.00} & 0.2843 {\footnotesize $\pm$ 0.00} & \snd{0.0533} {\footnotesize $\pm$ 0.00} &\snd{0.4461} {\footnotesize $\pm$ 0.01} & 0.0499 {\footnotesize $\pm$ 0.00} & 0.4228 {\footnotesize $\pm$ 0.01} & 0.0408 {\footnotesize $\pm$ 0.00} & 0.3710 {\footnotesize $\pm$ 0.01} & 0.0464 {\footnotesize $\pm$ 0.00} & 0.4450 {\footnotesize $\pm$ 0.01} \\
\hline
\model & \fst{0.0671} {\footnotesize $\pm$ 0.01} & \fst{0.7028} {\footnotesize $\pm$ 0.08} & \fst{0.0704} {\footnotesize $\pm$ 0.01} & \fst{0.5621} {\footnotesize $\pm$ 0.07} & \fst{0.0832} {\footnotesize $\pm$ 0.01} & \fst{0.7176} {\footnotesize $\pm$ 0.13} & \fst{0.0786} {\footnotesize $\pm$ 0.00} & \fst{0.7262} {\footnotesize $\pm$ 0.07} & \fst{0.0560} {\footnotesize $\pm$ 0.00} & \fst{0.6108} {\footnotesize $\pm$ 0.11} & \fst{0.0625} {\footnotesize $\pm$ 0.01} & \fst{0.6368} {\footnotesize $\pm$ 0.08} \\
\textbf{Gain (\%)} & \textbf{22.45\%} & \textbf{$\geq$100\%} & \textbf{31.84\%} & \textbf{85.08\%} & \textbf{56.10\%} & \textbf{60.86\%} & \textbf{47.47\%} & \textbf{64.93\%} & \textbf{33.65\%} & \textbf{47.79\%} & \textbf{10.62\%} & \textbf{8.87\%}\\

\textbf{p-value} & \textbf{6.9$\cdot$10$^{-3}$} & \textbf{3.0$\cdot$10$^{-5}$} & \textbf{1.7$\cdot$10$^{-4}$} & \textbf{3.0$\cdot$10$^{-6}$} & \textbf{1.8e$\cdot$10$^{-5}$} & \textbf{4.3$\cdot$10$^{-4}$} & \textbf{4.6e$\cdot$10$^{-6}$}& \textbf{2.1e$\cdot$10$^{-5}$}& \textbf{4.7e$\cdot$10$^{-5}$}& \textbf{3.1e$\cdot$10$^{-4}$} & \textbf{6.3e$\cdot$10$^{-3}$}& \textbf{ 7.9e$\cdot$10$^{-3}$} \\
\specialrule{.1em}{0em}{0em}
\end{tabular}
}
\vspace{-0.2cm}
\end{table*}

\begin{table}[t]

\centering
\caption{Data statistics for public benchmarking evaluation.}
\label{tb:publicData}
\resizebox{\linewidth}{!}{
\begin{tabular}{c|ccc|ccc}
\specialrule{.1em}{0em}{0em}
\multirow{2}{*}{Asset}  &\multicolumn{3}{c|}{Training} &\multicolumn{3}{c}{Evaluation} \\
~ & \#Ins. & Time & \#Trans. & \#Ins. & Time & \#Trans.\\
\hline 
\textbf{C-futures}     & 2,770    & 08Q1-16Q4   & 2,190 & 2,243& 17Q1-20Q3 & 871 \\
\textbf{F-futures}     & 424     & 08Q1-16Q4   & 2,190 & 344& 17Q1-20Q3 & 871\\
\cline{1-7}
\textbf{CSI100}     & 100   & 08Q1-16Q4   & 2,190 & 100& 17Q1-20Q3 & 871 \\
\textbf{CSI300}     & 300  & 08Q1-16Q4   & 2,190 & 300& 17Q1-20Q3 & 871\\
\textbf{CSI500}     & 500   & 08Q1-16Q4   & 2,190 & 500& 17Q1-20Q3 & 871\\
\specialrule{.1em}{0em}{0em}
\end{tabular}}
\end{table}

\subsection{Public Benchmarking Evaluation (RQ2)}

\subsubsection{\textbf{Evaluation Data}}
For a comprehensive benchmarking with existing models, we also include CSI100 index data for comparison, in addition to the two most widely studied indexes, CSI300 and CSI500.
Data statistics are reported in Table~\ref{tb:publicData}.

\subsubsection{\textbf{Competing Methods.}}
We compare \model~ with fourteen representative models of stock price forecasting, i.e., deep-neural-network-based and decision-tree-based methods.
Due to page limits, we attach their detailed introduction in~\cref{app:methods}.
\begin{itemize}[leftmargin=*]
\item \textbf{Deep-neural-network-based (DNN-based) methods.} 
These approach analyzes sequential time steps through diverse neural architectures to forecast prices, encompassing nine methods:
LSTM~\cite{lstm}, GRU~\cite{gru}, Transformer~\cite{vaswani2017attention}, ALSTM~\cite{qin_dual-stage_2017}, SFM~\cite{sfm},  TCN~\cite{tcn}, TabNet~\cite{tabnet}, TFT~\cite{tft}, TRA~\cite{lin2021learning} 
and Localformer~\cite{localformer}.

\item \textbf{Decision-tree-based methods.} 
These approach sequentially builds decision trees where successive models focus on correcting predecessor errors, maintaining strong predictive power as classic trading models. The included representatives are XGBoost~\cite{chen2016xgboost}, CatBoost~\cite{catboost}, LightGBM~\cite{ke2017lightgbm}, and DoubleEnsemble~\cite{zhang2020doubleensemble}.
\end{itemize}

\subsubsection{\textbf{Evaluation Metrics.}}
We follow
~\cite{qlib} to include four commonly used metrics:
(1) Information Coefficient (IC), (2) Rank Information Coefficient (RIC), (3) Information Ratio of IC (IR$_{\text{IC}}$), and (4) Information Ratio of Rank IC (IR$_{\text{Rank IC}}$). 
Higher values indicate better performance.
Explanations are detailed in~\cref{app:metrics}.

\subsubsection{\textbf{Overall Performance Analysis}}
We report the five-time averaged results in Table~\ref{tab:public_performance} with the following discussions:
\begin{itemize}[leftmargin=*]
\item DNN-based methods show relatively competitive performance but some of them rank lower than decision-tree-based methods, particularly in IC and IR$_{\text{IC}}$ metrics. 
Decision-tree-based methods, e.g., DoubleEnsemble, perform well in RIC and IR$_\text{rank IC}$, with some instances highlighted, suggesting that they are robust in ranking-related tasks.
Particularly, TRA and DoubleEnsemble are representative models with outstanding performance. 

\item \model~ consistently achieves the best performance, as shown by the highest metric values, i.e., IC, IR$_{\text{IC}}$, RIC, IR$_\text{rank IC}$. 
Furthermore, the performance gains are substantial, with improvements from 8.87\% to 139.62\% across different data and metrics. 
We also conduct the paired t-tests.
Each associated p-value also confirms that the performance improvements of our model over others are statistically significant.

\item Furthermore, \model~performs significantly better on CSI100.
This is because CSI100 represents the Top-100 most stable blue-chip stocks and \model~can more easily capture patterns and dependencies among them and market futures. For larger stock pools CSI300 and CSI500, the greater stock diversity introduces increased variability and complexity. 
And our model is still competitive, indicating its adaptability across different stock pool sizes in practical deployment.
\end{itemize}

\subsection{Empirical Analyses of \model~(RQ3)}
\subsubsection{\textbf{Ablation Study.}}
To study the effect of each proposed module, we introduce several model variants by disabling their functionality. We report the results of experimenting on CSI300 data in Table~\ref{tb:ablation} and provide the analyses accordingly.
\begin{enumerate}[leftmargin=*]
    \item \underline{w/o futures information}: This variant disables the encoding of commodity and financial futures information aggregation in \text{CME-Layer}, relying solely on the stock features. 
    The performance drops significantly across all metrics, ranging from 8.78\% to 24.76\%, highlighting the importance of futures information synergy in stock price forecasting.

    \item \underline{w/o \texttt{BDP-Former}}: This variant replaces our entire \texttt{BDP-Former} by first adding futures and stock features and then passing them through a two-layer MLP before predicting final prices.
    The performance gap between this variant and our \model~is substantial, demonstrating the effectiveness of \texttt{BDP-Former} in capturing latent correlations between diverse futures and stocks for price movement forecasting.

    \item \underline{w/o pseudo-volatility}: Lastly, we remove pseudo-volatility estimation, and replace the associated volatility-aware optimization with the MSE objective. Our experimental results confirm that the integration of such pseudo-volatility optimization is crucial for improving prediction accuracy.
    We provide a more detailed empirical analysis of this learning paradigm in \cref{exp:loss}. 
\end{enumerate}

\subsubsection{\textbf{Pseudo-volatility Compatibility on Existing Models.}}
We also explore the compatibility of our pseudo-volatility (PV) design with other DNN-based models, e.g., LocalFormer~\cite{localformer} and TRA~\cite{lin2021learning}.
Specifically, we retain their original feature encoding and decoding parts but incorporate our \texttt{PV-Estimator} and adjust their loss function accordingly. 
As shown in Table~\ref{tab:pv}, both PV-integrated models demonstrate notable performance improvements, verifying the design effectiveness of our pseudo-volatility optimization.

\begin{table}[t]
\centering
\caption{Ablation study on CSI300 data.}
\label{tb:ablation}
\tabcolsep=0.3cm
\resizebox{\linewidth}{!}{
\begin{tabular}{c|cccc}
\specialrule{.1em}{0em}{0em}
\textbf{Variant} &\textbf{IC} & \textbf{IR$_{\text{IC}}$} & \textbf{RIC} & \textbf{IR$_{\text{Rank IC}}$} \\
\cline{1-5}
\multirow{2}{*}{w/o futures information} & 0.0626 & 0.6546 & 0.0627& 0.6263\\
~ &\drp{-24.76\%} &\drp{-8.78\%} &\drp{-20.23\%} &\drp{-13.76\%} \\
\cline{1-5}
\multirow{2}{*}{w/o \texttt{BDP-Former}} & 0.0570 &0.3179 & 0.0514& 0.2722\\
~  &\drp{-31.49\%} &\drp{-55.70\%} &\drp{-34.61\%} &\drp{-62.52\%} \\
\cline{1-5}
\multirow{2}{*}{w/o pseudo-volatility} & 0.0645 &0.6465 &0.0647 & 0.6344\\
~  &\drp{-22.48\%} &\drp{-9.91\%} &\drp{-17.68\%} &\drp{-12.64\%} \\
\cline{1-5}
\model & \textbf{0.0832}  & \textbf{0.7176}  & \textbf{0.0786}  & \textbf{0.7262}\\
\specialrule{.1em}{0em}{0em}
\end{tabular}
}
\vspace{-0.3cm}
\end{table}

\begin{table}[t]
    \centering
    \caption{Performance of pseudo-volatility-based models.}
    \label{tab:pv}
    \tabcolsep=0.1cm
    \resizebox{\linewidth}{!}{
    \begin{tabular}{c|cccc}
    \specialrule{.1em}{0em}{0em}
        \textbf{Method} & \textbf{IC} & \textbf{IR$_{\text{IC}}$} & \textbf{RIC} & \textbf{IR$_{\text{Rank IC}}$} \\
        \cline{1-5}
        LocalFormer & 0.0356 & 0.2756 & 0.0468 & 0.3784 \\
        \rowcolor{gap} LocalFormer$_{PV}$ & \textbf{0.0415} {\footnotesize(+16.57\%)} & \textbf{0.3165} {\footnotesize(+14.84\%)} & \textbf{0.0508} {\footnotesize(+8.55\%)}  & \textbf{0.4114} {\footnotesize(+8.72\%)} \\
        TRA & 0.0440 & 0.3535 & 0.0540 & 0.4451 \\
        \rowcolor{gap} TRA$_{PV}$ & \textbf{0.0520} {\footnotesize(+18.18\%)} & \textbf{0.4243} {\footnotesize(+20.03\%)} & \textbf{0.0553} {\footnotesize(+2.41\%)} & \textbf{0.4821} {\footnotesize(+8.31\%)} \\
        \specialrule{.1em}{0em}{0em}
    \end{tabular}}
    \vspace{-0.3cm}
\end{table}

\begin{figure}[t]
    \hspace{-1em}
    \begin{minipage}{0.24\textwidth}
        \includegraphics[width=\textwidth]{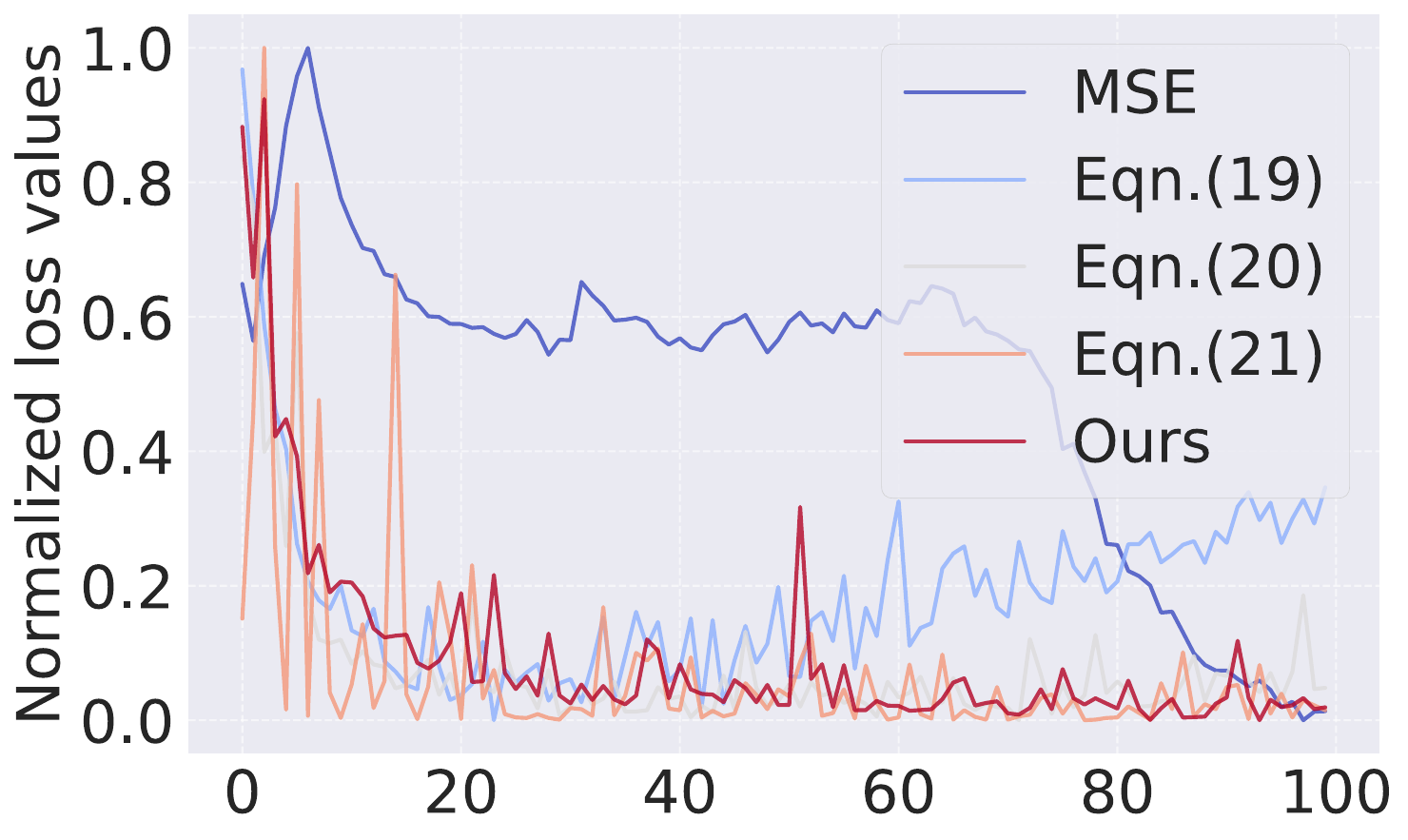}
        \vspace{-1em}
    \end{minipage}    
    \begin{minipage}{0.24\textwidth}
        \includegraphics[width=\textwidth]{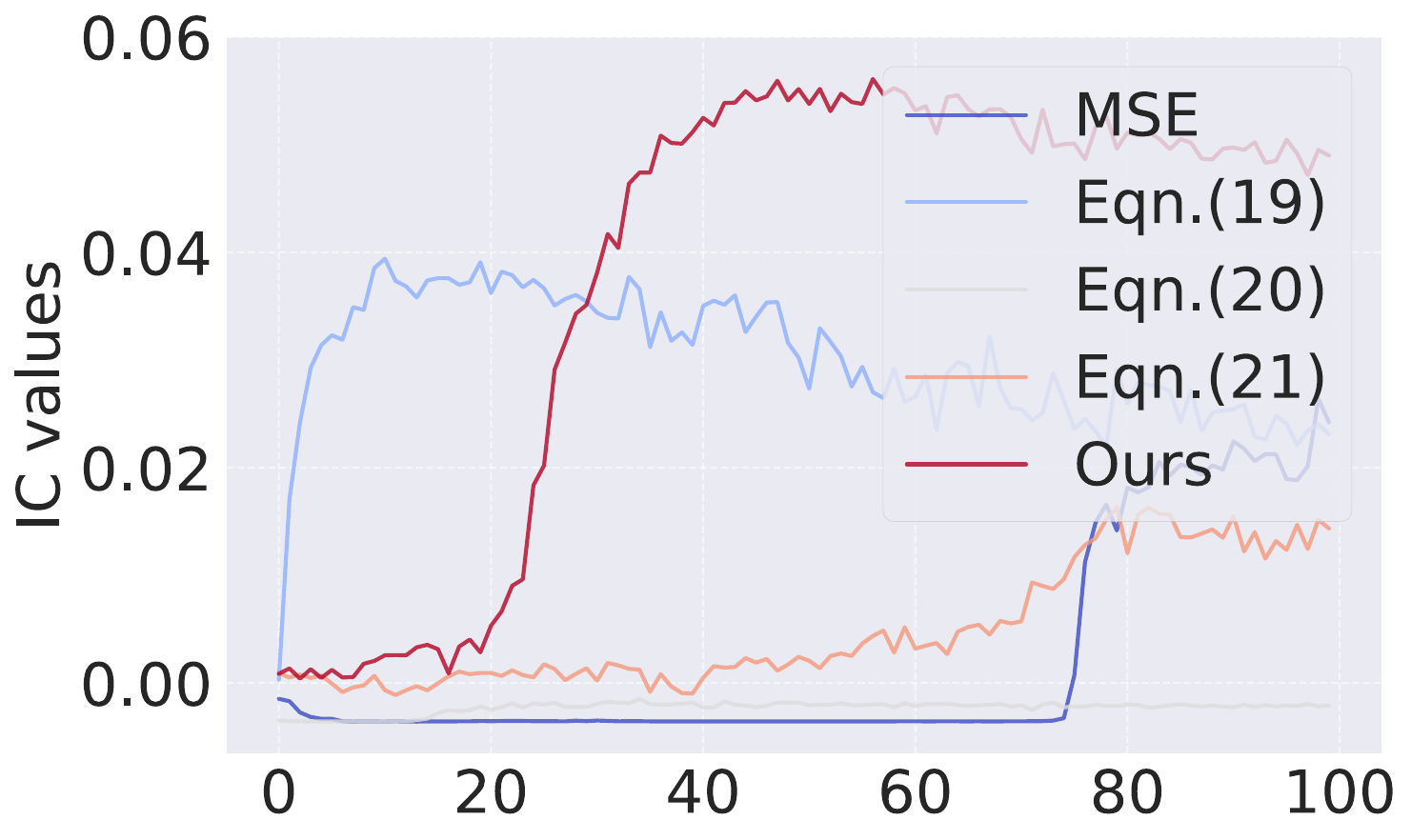}
        \vspace{-1em}
    \end{minipage}
    \caption{Empirical comparison of objective functions.}
    \label{fig:loss_curve}
    \vspace{-0.2cm}
\end{figure}

\subsubsection{\textbf{Evaluation of Objective Function Designs.}}
\label{exp:loss}

Lastly, in Figure~\ref{fig:loss_curve}, we compared our final objective function of Eqn.~(\ref{eq:final}), with MSE and other functions of Eqn.'s (\ref{eq:hete_mse})$\sim$(\ref{eq:regular}) on 100 epochs training. 
For the left-hand side figure of loss values, we apply the min-max normalization for value scaling between [0,1]. 
For the right-hand side figure, we report corresponding IC values.
We observe that, compared to other loss designs, while our final objective and Eqn.~(\ref{eq:regular}) present better convergence, our objective shows a more stable training process and superior IC value performance.
\section{Related Work}
\label{sec:related}

\textbf{Stock Price Movement Forecasting.} 
Stock price movement forecasting (SPMF) has been a longstanding challenge in the financial domain due to the temporal and non-linear complexities inherent in financial data. 
Traditional approaches primarily rely on fundamental analysis, utilizing manually engineered features and macroeconomic indicators \cite{box1970distribution,holt2004forecasting,kavitha2013stock}. 
The advent of machine learning introduced models such as decision trees \cite{quinlan1986induction,quinlan2014c4} and gradient boosting trees (GBTs) \cite{chen2016xgboost,ke2017lightgbm} improves predictive performance by capturing the dynamic and nonlinear nature of market behavior~\cite{nayak2015naive,zhang2020doubleensemble}. 
Deep learning models have recently revolutionized SPMF by enabling the direct utilization of raw time-series data, reducing reliance on manually engineered features.
Specifically, RNN-based methods \cite{zhang2017stock,li2018stock,wang2022adaptive} have shown success in modeling temporal dependencies; 
CNN-based methods~\cite{hoseinzade2019cnnpred,lu2021cnn} treat historical price data as structured input, effectively extracting localized patterns.
Graph-based approaches \cite{li2021modeling,sawhney2021stock} studies the intricate interdependence among different stocks, which is a a common methodology in various applications~\cite{cstar,chen2023star}. 
Furthermore, recent works have incorporated additional data sources with graph structure modeling~\cite{zeng2025efficient,zeng2021efficient}, such as exchanges~\cite{cao2012multifractal}, sales~\cite{zhang2022co}, and earnings calls~\cite{medya2022exploratory,yuan2023earnings,liu2024echo}, for information enhancement~\cite{hiang2005cross,cao2012multifractal}.
Some other efforts try to adjust and stabilize model predictions with consideration of stock price volatility~\cite{duan2022factorvae,zhu2022wise,li2024finreport}.

\textbf{Transformer-based Models for SPMF.} 
Transformer structures~\cite{vaswani2017attention,kenton2019bert} are widely used in many applications~\cite{chen2024deep,lin2024effective,chen2022effective,chen2020literature,wu2023survey,qiu2024ease}. 
It further has emerged as a leading approach in financial time-series forecasting~\cite{localformer,zhou2021informer,wu2021autoformer,nie2022time,chen2024pathformer}. 
Recent studies have introduced various enhancements to Transformer architectures. 
For instance, methods~\cite{ding2020hierarchical,wang2022adaptive} capture multi-scale financial dependencies through enhanced feature extraction, showing strong prediction performance across markets.
Other works have explored the integration of external data sources to enhance prediction. 
TEANet~\cite{zhang2022transformer} fuses social media text and prices for temporal modeling. 
StockFormer~\cite{gao_stockformer_2023} adopts a hybrid approach by integrating predictive coding with reinforcement learning. 
Meanwhile, MASTER~\cite{li2024master} addresses cross-time and momentary stock correlation through market-guided feature selection. 
Generally, these transformer-based models capture long-range dependencies and complex interactions and have demonstrated remarkable performance superiority in stock price forecasting.

\section{Conclusions and Future Work}
\label{sec:con}
We propose \model, featuring the \texttt{BDP-Former} architecture that jointly models temporal patterns from futures/stock markets and price inter-correlations. The framework introduces pseudo-volatility guided loss weighting to enhance stability under real trading conditions. Extensive evaluations demonstrate superiority in both industrial backtesting and academic benchmarks.
For future work, two directions emerge as critical: (1) Integrating rigorously filtered LLM insights~\cite{zhao2024revolutionizing,li2024llms,liu2024water,li2023vision} and large-scale graph data management~\cite{zeng2024distributed,zeng2025Skyline} to enhance cross-market analysis while ensuring information credibility; (2) Developing continual learning protocols~\cite{zhang2024influential,yu2024recent} for efficient model adaptation to streaming financial data while maintaining prediction accuracy.

\begin{acks}
This work was partially supported by NSFC under Grant 62302421, Basic and Applied Basic Research Fund in Guangdong Province under Grant 2023A1515011280, Ant Group through CCF-Ant Research Fund, Shenzhen Research Institute of Big Data under grant SIF20240004, and the Guangdong Provincial Key Laboratory of Big Data Computing, The Chinese University of Hong Kong, Shenzhen.
\end{acks}

\bibliographystyle{ACM-Reference-Format}
\bibliography{reference}
\appendix

\section{Experimental Details}

\subsection{Evaluation Metrics}
\label{app:metrics}

\begin{itemize}[leftmargin=*]
    \item \textbf{Annualized Return (AR)}: AR is defined as the annualized rate of return on an investment. 
    
    \item \textbf{Winning Rate (WR)}: WR is defined as the percentage of profitable trades relative to the total trades executed.
    
    \item \textbf{Sharpe Ratio (ShR)}: ShR measures the risk-adjusted return of an investment by comparing the return to its standard deviation. A higher ratio signifies better risk-adjusted performance.

    \item \textbf{Sortino Ratio (SoR)}: Similar to ShR, SoR evaluates risk-adjusted returns uses only downside deviation. This metric focuses on minimizing losses.

    \item \textbf{Maximum Drawdown (MD)}: MD reflects the largest peak-to-trough decline over a specified period. It measures the extent of the worst loss sustained before recovery.

    \item \textbf{Maximum Drawdown Duration (MD-D)}: MD-D indicates the time taken for an investment to recover from its maximum drawdown to its previous peak.

    \item \textbf{Turnover Rate (TR)}: TR proportion of assets replaced with which assets within a portfolio are traded over a specified period. 

    \item \textbf{Information Coefficient (IC)}: 
    IC is a key metric in financial analysis that measures the correlation between predictions and realized returns.
    
    \item \textbf{Rank IC (RIC)}: RIC is similar to IC but is computed using rank-based correlation.
    It is useful for long-short strategy which rely on the model's ranking capacity.
    
    \item \textbf{Information Ratio of IC (IR$_{\text{IC}}$)}:
    IR$_{\text{IC}}$ is the ratio of the IC to its standard deviation, quantifying the stability of the model’s performance over time.
    \item \textbf{Information Ratio of Rank IC (IR$_{\text{Rank IC}}$)}: 
    The ratio of Rank IC to its standard deviation, reflecting the stability of the rank-based performance.
\end{itemize}

\subsection{\textbf{Evaluation Configurations}}
\label{app:setup}
We implement using Python 3.8 and PyTorch 1.13.1 with non-distributed training.
The experiments are run on a Linux machine with 8 NVIDIA A100 GPUs and 6 Intel(R) Xeon(R) Platinum 8350C CPUs with 2.60GHz.
We adhere to all baselines' officially reported hyper-parameter settings and conduct a grid search for models without prescribed configurations.
For a fair comparison, we fix the embedding dimension at 512. 
The learning rate is tuned in the range \{$10^{-5}, 10^{-4}, 10^{-3}$\}. 
Optimization for all models is performed using the default AdamW optimizer~\cite{adam}.

\subsection{Details of Competing Methods}
\label{app:methods}

\begin{itemize}[leftmargin=*]
\item \textbf{Deep-neural-network-based (DNN-based) methods.}  
    \begin{enumerate}[leftmargin=10pt]
    \item \underline{LSTM~\cite{lstm}} captures long-term dependencies through memory cells, effectively modeling sequential data.
    \item \underline{GRU~\cite{gru}} uses gating mechanisms to capture dependencies in sequential data with fewer parameters than LSTM.
    \item \underline{Transformer~\cite{vaswani2017attention}} utilizes self-attention to capture long-range dependencies, suitable for time series forecasting.
    \item \underline{ALSTM~\cite{qin_dual-stage_2017}} integrates attention mechanisms into LSTM, selectively focusing on relevant time steps.
    \item \underline{SFM~\cite{sfm}} captures temporal features by encoding state and frequency representations.
     \item \underline{TCN~\cite{tcn}} employs causal convolutions and flexible receptive fields for modeling sequential data.
    \item \underline{TabNet~\cite{tabnet}} dynamically selects features with sequential attention for effective tabular data modeling.
    \item \underline{Localformer~\cite{localformer}} enhances transformers by focusing on local temporal dynamics.
    
    \item \underline{TRA~\cite{lin2021learning}} applies Transformer-based relational attention to capture long-term dependencies in graph-structured data.
    \end{enumerate}

\item \textbf{Gradient-boosting-based methods.} 
    \begin{enumerate}[leftmargin=10pt,resume]
    \item \underline{XGBoost~\cite{chen2016xgboost}} proposes a sparsity-aware algorithm and weighted quantile sketch for approximate tree learning.

    \item \underline{CatBoost~\cite{catboost}} efficiently handles categorical features using ordered boosting to improve the performance. 
    
    \item \underline{LightGBM~\cite{ke2017lightgbm}} employs histogram-based methods and leaf-wise tree growth for better scalability and prediction accuracy.
    
    \item \underline{DoubleEnsemble~\cite{zhang2020doubleensemble}} combines two ensemble strategies to improve generalization, particularly for imbalanced data.
    \end{enumerate}
\end{itemize}

\end{document}